\documentclass[journal]{IEEEtran}
\ifCLASSINFOpdf
  \usepackage[pdftex]{graphicx}
\else
\fi
\usepackage{amsmath,amssymb}
\ifCLASSOPTIONcompsoc
 \usepackage[caption=false,font=normalsize,labelfont=sf,textfont=sf]{subfig}
\else
 \usepackage[caption=false,font=footnotesize]{subfig}
\fi
\usepackage{algorithm}
\usepackage[noend]{algpseudocode}
\usepackage{bm}
\usepackage{booktabs}

\DeclareMathOperator*{\argmax}{arg\,max}
 
\DeclareMathOperator{\sgn}{sgn}

\newcommand{\appropto}{\mathrel{\vcenter{
  \offinterlineskip\halign{\hfil$##$\cr
    \propto\cr\noalign{\kern2pt}\sim\cr\noalign{\kern-2pt}}}}}

\newcommand{\minus}{\scalebox{0.75}[1.0]{$-$}}

\begin{document}
%
\title{Combining Planning and Deep Reinforcement Learning in Tactical Decision Making for Autonomous Driving}
%
%
%

\author{Carl-Johan~Hoel,~Katherine~Driggs-Campbell,~Krister~Wolff,~Leo~Laine,~and~Mykel~J.~Kochenderfer%
\thanks{
This work was partially supported by the Wallenberg Artificial Intelligence, Autonomous Systems, and Software Program (WASP), funded by the Knut and Alice Wallenberg Foundation, and partially by Vinnova FFI. \textit{(Corresponding author: Carl-Johan Hoel.)}}%
\thanks{C. J. Hoel and L. Laine are with the Department of Vehicle Automation, Volvo Group, 40508 Gothenburg, Sweden, and with the Department of Mechanics and Maritime Sciences,
Chalmers University of Technology, 41296 Gothenburg, Sweden (\mbox{e-mail}: \{carl-johan.hoel, leo.laine\}@volvo.com). }%
\thanks{K. Driggs-Campbell is with the Electrical and Computer Engineering
Department, University of Illinois at Urbana-Champaign, IL 61801, USA (\mbox{e-mail}: krdc@illinois.edu) }%
\thanks{K. Wolff is with the Department of Mechanics and Maritime Sciences,
Chalmers University of Technology, 41296 Gothenburg, Sweden (\mbox{e-mail}: \mbox{krister.wolff@chalmers.se})}%
\thanks{M. J. Kochenderfer is with the Aeronautics and Astronautics Department at Stanford University, Stanford, CA 94305, USA (\mbox{e-mail}: mykel@standford.edu). }%
}

\maketitle


\begin{abstract}

Tactical decision making for autonomous driving is challenging due to the diversity of environments, the uncertainty in the sensor information, and the complex interaction with other road users. This paper introduces a general framework for tactical decision making, which combines the concepts of planning and learning, in the form of Monte Carlo tree search and deep reinforcement learning. The method is based on the AlphaGo Zero algorithm, which is extended to a domain with a continuous state space where self-play cannot be used. The framework is applied to two different highway driving cases in a simulated environment and it is shown to perform better than a commonly used baseline method. The strength of combining planning and learning is also illustrated by a comparison to using the Monte Carlo tree search or the neural network policy separately.

\end{abstract}


\begin{IEEEkeywords}
Autonomous driving, tactical decision making, reinforcement learning, Monte Carlo tree search.
\end{IEEEkeywords}

%
\IEEEpeerreviewmaketitle


\section{Introduction}
\label{sec:introduction}
%
%
%
%

\IEEEPARstart{A}{utonomous} vehicles are expected to bring many societal benefits, such as increased productivity, a reduction of accidents, and better energy efficiency~\cite{FAGNANT2015167}.
One of the technical challenges for autonomous driving is to be able to make safe and effective decisions in diverse and complex environments, based on uncertain sensor information, while interacting with other traffic participants.
A decision making method should therefore be sufficiently general to handle the spectrum of relevant environments and situations. The naive approach of trying to anticipate all possible traffic situations and manually code a suitable behavior for these is infeasible.
The topic of this paper is tactical decision making, which considers high level decisions that adapts the behavior of the vehicle to the current traffic situation~\cite{Michon1985}. For example, these decisions could handle when to change lanes, or whether or not to stop at an intersection.

Rule-based methods, implemented as handcrafted state machines, were successfully used during the DARPA Urban Challenge~\cite{darpaCMU},~\cite{darpaStanford},~\cite{darpaVirginia}. However, a drawback with these rule-based approaches is that they lack the ability to generalize to unknown situations, which makes it hard to scale them to the complexity of real-world driving.
Another approach is to treat the decision making task as a motion planning problem, which has been applied to highway~\cite{Werling2010},~\cite{Nilsson2015} and intersection scenarios~\cite{Damerow2015}. While successful for some situations, the sequential design of this method, which first predicts the trajectory of the surrounding vehicles and then plan the ego vehicle trajectory accordingly, results in a reactive behavior which does not consider interaction during the trajectory planning.

It is common to model the tactical decision making task as a partially observable Markov decision process (POMDP)~\cite{Kochenderfer2015}. This mathematical framework models uncertainty in the current state, future evolution of the traffic scene, and interactive behavior.
The task of finding the optimal policy for a POMDP is difficult, but many approximate methods exist.
Offline methods can solve complex situations and precomputes a policy before execution, which for example has been done for an intersection scenario~\cite{Brechtel2014},~\cite{Bai2014}. However, due to the large number of possible real world scenarios, it becomes impossible to precalculate a policy that is generally valid.
Online methods compute a policy during execution, which makes them more versatile than offline methods, but the limited computational resources reduces the solution quality. Ulbrich et al.~considered a lane changing scenario on a highway, where they introduced a problem-specific high level state space that allowed an exhaustive search to be performed~\cite{Ulbrich2015}. Another online method for solving a POMDP is Monte Carlo tree search (MCTS)~\cite{Browne2012}, which has been applied to lane changes on a highway~\cite{Sunberg2017}. Hybrid approaches between offline and online planning have also been studied~\cite{Sonu2018}.

Reinforcement learning (RL)~\cite{Sutton2018} has proved successful in various domains, such as playing Atari games~\cite{Mnih2015}, in continuous control~\cite{Lillicrap2015}, and reaching a super human performance in the game of Go~\cite{Silver2017}.
In a previous paper, we showed how a Deep Q-Network (DQN) agent could learn to make tactical decisions in two different highway scenarios~\cite{Paper2}. 
A similar approach, but applied to an intersection scenario, was presented by Tram et al.~\cite{Tram2018}.
Expert knowledge can be used to restrict certain actions, which has been shown to speed up the training process for a lane changing scenario~\cite{Mukadam2017}. A different approach, which uses a policy gradient method to learn a desired behavior, has been applied to a merging scenario on a highway~\cite{Shalev2016}.
A common drawback with RL methods is that they require many training samples to reach convergence. RL methods may also suffer from the credit assignment problem, which makes it hard to learn long temporal correlations between decisions and the overall performance of the agent~\cite{Sutton2018}.

This paper presents a general framework, based on the AlphaGo Zero algorithm~\cite{Silver2017}, that combines the concepts of planning and learning to create a tactical decision making agent for autonomous driving (Sect.~\ref{sec:approach}). The planning is done with a variation of Monte Carlo tree search, which constructs a search tree based on random sampling.
The difference between standard MCTS and the version used here is that a neural network biases the sampling towards the most relevant parts of the search tree.
The neural network is trained by a reinforcement learning algorithm, where the MCTS component both reduces the required number of training samples and aids in finding long temporal correlations. 
The presented framework is applied to two conceptually different driving cases (Sect.~\ref{sec:implementation}), and performs better than a commonly used baseline method (Sect.~\ref{sec:results}). To illustrate the strength of combining planning and learning, it is compared to using the planning or the learned policy separately.

In contrast to the related work, the approach that is introduced in this paper combines the properties of planning and RL. When used online, the planning can be interrupted anytime with a reasonable decision, even after just one iteration, which will then return the learned action. More computational time will improve the result.
The proposed approach is general and can be adapted to different driving scenarios. Expert knowledge is used to ensure safety by restricting actions that lead to collisions. The intentions of other vehicles are considered when predicting the future and the algorithm operates on a continuous state space.
The AlphaGo Zero algorithm is here extended beyond the zero-sum board game domain of Go, to a domain with a continuous state space, a not directly observable state, and where self-play cannot be used.
Further properties of the framework are discussed in Sect.~\ref{sec:discussion}.

The main contributions of this paper are:
\begin{itemize}
    \item The extension of the AlphaGo Zero algorithm, which allows it to be used in the autonomous driving domain.
    \item The introduction of a general tactical decision making framework for autonomous driving, based on this extended algorithm.
    \item The performance analysis of the introduced tactical decision making framework, applied to two different test cases.
\end{itemize}


\section{Theoretical Background}
\label{sec:theoreticalBackground}

This section introduces partially observable Markov decision processes and two solution methods: Monte Carlo tree search, and reinforcement learning. 

\subsection{Partially Observable Markov Decision Process}
\label{sec:decisionMaking}

A POMDP is defined as the tuple $( \mathcal{S}, \mathcal{A}, \mathcal{O}, T, O, R, \gamma )$, where $\mathcal{S}$ is the state space, $\mathcal{A}$ is the action space, $\mathcal{O}$ is the observation space, $T$ is a state transition model, $O$ is an observation model, $R$ is a reward model, and $\gamma$ is a discount factor. At every time step $t$, the goal of the agent is to maximize the future discounted return, defined as
\begin{align}
    \label{eq:discountedReturn}
    R_t = \sum_{k=0}^\infty \gamma^k r_{t+k},
\end{align}
where $r_{t+k}$ is the reward at step $t+k$~\cite{Kochenderfer2015}.

The state transition model $T(s' \mid s,a)$ describes the probability that the system transitions to state $s'$ from state $s$ when action $a$ is taken. The observation model $O(o \mid s,a,s')$ is the probability of observing $o$ in state $s'$, after taking action $a$ in state $s$. For many real world problems, it is not feasible to represent the probability distributions $T$ and $O$ explicitly. However, some solution approaches only require samples and use a generative model $G$ instead, which generates a new sampled state and observation from a given state and action, i.e., $(s', o) \sim G(s,a)$. The reward model defines the reward of each step as $r=R(s,a,s')$~\cite{Kochenderfer2015}.

Since the agent cannot directly observe the state $s$ of the environment, it can maintain a belief state $b$, where a probability $b(s)$ is assigned to being in state $s$. In simple cases, the belief can be exactly calculated using Bayes' rule. However, approximate methods, such as particle filtering, are often used in practice~\cite{Thrun2005}.

\subsection{Monte Carlo Tree Search}
\label{sec:mcts}
MCTS can be used to select approximately optimal actions in POMDPs~\cite{Browne2012}. It constructs a search tree that consists of alternating layers of state nodes and action nodes, in order to estimate the state-action value function $Q(s,a)$, which describes the expected return $R_t$ when taking action $a$ in state $s$ and then following the given policy. A generative model $G$ is used to traverse the tree to the next state, given a state and an action. In the upper confidence tree version, the tree is expanded by selecting the action node that maximizes the upper confidence bound (UCB), defined as
\begin{align}
    UCB(s,a) = Q(s,a) + c_\mathrm{uct} \sqrt{\frac{\ln{N(s)}}{N(s,a)}},
\end{align}
where $Q(s,a)$ is the current estimate of the state-action value function, $N(s,a)$ is the number of times action $a$ has been tried from state $s$, $N(s)=\sum_{a\in\mathcal{A}}{N(s,a)}$, and $c_\mathrm{uct}$ is a parameter that controls the amount of exploration in the tree.

The standard formulation of MCTS cannot handle problems with a continuous state space, since then the same state may never be sampled more than once from $G$, which would result in a wide tree with just one layer. One way to address this problem is to use progressive widening~\cite{Couetoux2011}. With this technique, the number of children of a state-action node is limited to
$kN(s,a)^{\alpha}$,
where $k$ and $\alpha$ are tuning parameters. When there are less children than the limit, a new state is added by sampling from $G$. Otherwise, a previous child is randomly chosen. Thereby the number of children gradually grows as $N(s,a)$ increases.

\subsection{Reinforcement Learning}
\label{sec:reinforcementLearning}
Reinforcement learning is a branch of machine learning, where an agent tries to learn a policy $\pi$ that maximizes the expected future return $\mathbb{E}(R_t)$ in some environment~\cite{Sutton2018}. The policy defines which action $a$ to take in a given state $s$. In the RL setting, the state transition model $T$ (or $G$) of the POMDP may not be known. Instead, the agent gradually learns by taking actions in the environment and observing what happens, i.e., collecting experiences $(s,a,s',r)$.


\section{Tactical Decision Making Framework}
\label{sec:approach}

This paper introduces a framework that combines planning and learning for tactical decision making in the autonomous driving domain.
With this approach, a neural network is trained to guide MCTS to the relevant regions of the search tree, and at the same time, MCTS is used to improve the training process of the neural network.
This idea is based on the AlphaGo Zero algorithm, originally developed for the game of Go~\cite{Silver2017}.
However, such a zero-sum board game domain has several properties that cannot be used in a more general domain, such as a discrete state, symmetry properties, and the possibility of using self-play.
This section shows how the AlphaGo Zero algorithm was generalized to a domain with a continuous state space and where self-play cannot be used.

\subsection{Tree search}

A neural network $f_\theta$, with parameters $\theta$, is used to guide the MCTS. The network takes a state $s$ as input, and outputs the estimated value $V(s,\theta)$ of this state and a vector that represents the prior probabilities $\bm{p}(s,\theta)$ of taking different actions,
\begin{align}
\label{eq:ftheta}
    \left(\bm{p}(s,\theta),V(s,\theta)\right) = f_\theta(s).     
\end{align}
For $m_\mathrm{act}$ possible actions, $P(s,a_i,\theta)$ represents the prior probability of taking action $a_i$ in state $s$, i.e., $\bm{p}(s,\theta) = (P(s,a_1,\theta), \dots, P(s,a_{m_\mathrm{act}},\theta))$.

In order to select which action to take from a given state, the \Call{SelectAction}{} function from Algorithm~\ref{alg:treeSearch} is used. This function constructs a search tree, where each state-action node stores a set of statistics $\{N(s,a), Q(s,a), C(s,a)\}$, where $N(s,a)$ is the number of visits of the node, $Q(s,a)$ is the estimated state-action value, and $C(s,a)$ it the set of child nodes.
To build the search tree, $n$ iterations are done, where each iteration starts in the root node $s_0$ and continues for time steps $t=0,1,\dots,L$ until it reaches a leaf node $s_L$ at step $L$. At each step, the algorithm selects the action that maximizes the UCB condition
\begin{align}
    \label{eq:ucb}
    UCB(s,a,\theta) & = \frac{Q(s,a)}{Q_\mathrm{max}} \nonumber \\
    & + c_\mathrm{puct} P(s,a,\theta)  \frac{\sqrt{\sum_b{N(s,b)}+1}}{N(s,a)+1}.
\end{align}
Here, $c_\mathrm{puct}$ is a parameter that controls the amount of exploration in the tree. In order to keep $c_\mathrm{puct}$ constant over environments, the $Q$-values are normalized by $Q_\mathrm{max} = r_\mathrm{max}/(1-\gamma)$, where $r_\mathrm{max}$ is the maximum possible reward in one time step. The reward is also typically normalized, which then means that $r_\mathrm{max}=1$.
In order to perform additional exploration during the training phase (not during evaluation), Dirichlet noise is added to the prior probabilities. Therefore, during training, $P(s,a,\theta)$ is replaced with $(1-\varepsilon)P(s,a,\theta) + \varepsilon \eta$, where $\eta \sim \mathrm{Dir}(\beta)$, and $\varepsilon$ and $\beta$ are parameters that control the amount of noise.

When an action has been chosen, the progressive widening criterion is checked to decide whether a new child node should be expanded. If the number of children is larger than $kN(s,a)^\alpha$, a previous child node is uniformly sampled from the set $C(s,a)$, and the iteration continues down the search tree. However, if $|C(s,a)|\leq kN(s,a)^\alpha$, a new leaf node is created. The state of this leaf node $s_L$ is sampled from the generative model, $s_L \sim G(s_{L-1},a_{L-1})$, and the transition reward $r_{L-1}$ is given by the reward function $r_{L-1}$ = $R(s_{L-1},a_{L-1},s_L)$. The pair $(s_L,r_{L-1})$ is then added to the set of child nodes $C(s_{L-1},a_{L-1})$ and the estimated value of the node $V(s_L,\theta)$ is given by the neural network $f_\theta$. All the action nodes $\{a_*\}$ of the leaf state node are initialized such that $N(s_L,a_*)=0$, $Q(s_L,a_*)=V(s_L,\theta)$, and $C(s_L,a_*)=\emptyset $. 

Finally, the visit count $N(s,a)$ and $Q$-values $Q(s,a)$ of the parent nodes that were visited during the iteration are updated through a backwards pass.

After $n$ iterations, the tree search is completed and an action $a$ is sampled proportionally to the exponentiated visit count of actions of the root node $s_0$, according to
\begin{align}
\label{eq:policyVisitCount}
    \pi(a \mid s) = \frac{N(s,a)^{1/\tau}}{\sum_b{N(s,b)^{1/\tau}}},
\end{align}
where $\tau$ is a temperature parameter that controls the level of exploration. During the evaluation phase, $\tau \to 0$, which means that the most visited action is greedily chosen.

\begin{algorithm}[!t]

    \caption{Monte Carlo tree search, guided by a neural network policy and value estimate.}\label{alg:treeSearch}
    \begin{algorithmic}[1]
        \Function{SelectAction}{$s_0,n,\theta$}
            \For{$i \in 1:n$}
                \State \Call{Simulate}{$s_0,\theta$}
            \EndFor
            \State $\pi(a\mid s) \gets \frac{N(s,a)^{1/\tau}}{\sum_b{N(s,b)^{1/\tau}}}$
            \State $a \gets $ sample from $\pi$
            \State \Return $a, \pi$
        \EndFunction
        \Function{Simulate}{$s,\theta$}
        
            \If{$s$ is terminal}
                \State \Return 0
            \EndIf

            \State $a \gets \argmax_a  UCB(s,a,\theta) $

            \If{$|C(s,a)| \leq k N(s,a)^\alpha$} 
                \State $s' \sim G(s,a)$
                \State $r \gets R(s,a,s')$
                \State $C(s,a) \gets C(s,a) \cup \{(s',r)\}$
                \State $v \gets \begin{cases}
                                     0, & \text{\small if $s'$ is terminal}\\
                                     V(s',\theta), & \text{\small otherwise}
                                  \end{cases}$
                \State $q \gets r + \gamma v$
            \Else
                \State  $(s',r) \gets $ sample uniformly from $C(s,a)$
                \State $q \gets r + \gamma$\Call{Simulate}{$s',\theta$}
            \EndIf

            \State $N(s,a) \gets N(s,a) + 1$
            \State $Q(s,a) \gets Q(s,a) + \frac{q-Q(s,a)}{N(s,a)}$
            
            \State \Return $q$

        \EndFunction
    \end{algorithmic}
\end{algorithm}

\subsection{Training process}

Algorithm~\ref{alg:training} shows the process for generating training data, for optimizing the neural network parameters.
First, experiences from a simulated environment are generated.
For each new episode, a random initial state is sampled and then the episode is run until termination, at step $N_s$, with actions chosen according to the \Call{SelectAction}{} function of Algorithm~\ref{alg:treeSearch}.
Upon termination, the discounted return $z_i$ that was received during the episode is calculated for each step $i=0,\dots,N_s-1$ by summing the rewards $r_i$ of the episode, according to
\begin{align}
    z_i = \sum_{k=i}^{N_s-1}\gamma^{k-i}r_k + \gamma^{N_s-i}v_\mathrm{end}.
\end{align}
If the final state $s_{N_s}$ is a terminal state, its value is set to zero, i.e., $v_\mathrm{end}=0$. Otherwise, the value is estimated as $v_\mathrm{end}=V(s_{N_s},\theta)$.
For each of the states $s_i$, the received discounted return $z_i$ and the action distribution from the search tree,  $\bm{\pi}_i = (\pi(a_1 \mid s_i),\dots,\pi(m_\mathrm{act} \mid s_i))$, are used as targets for training the neural network. The tuples $(s_i,\bm{\pi}_i,z_i)$ are therefore added to a memory of experiences.

In parallel to the collection of new training samples, the neural network parameters $\theta$ are optimized from the stored samples by using a gradient descent algorithm~\cite{Ruder2016}. A loss function $\ell$ is calculated as the sum of the mean-squared value error, the cross entropy loss of the policy, and an $L_2$ weight regularization term,
\begin{align}
\label{eq:loss}
    \ell = c_1(z-V(s,\theta))^2 - c_2\bm{\pi}^\top \log{\bm{p}(s,\theta)} + c_3\|\theta\|^2,
\end{align}
where $c_1$, $c_2$, and $c_3$ are parameters that balance the different parts of the loss.

\begin{algorithm}[!t]
    \caption{Procedure for generating training data.}\label{alg:training}
    \begin{algorithmic}[1]
        \Function{GenerateTrainingData}{}
            \While{network not converged}
                \State $s_0 \gets $ \Call{GenerateRandomState}{}
                \State $i \gets 0$
                \While{episode not finished}
                    \State $a_i, \pi_i \gets $ \Call{SelectAction}{$s_i,n,\theta$}
                    \State $s_{i+1}, r_i \gets $ \Call{StepEnvironment}{$s_i,a$}
                    \State $i \gets i + 1$
                \EndWhile
                \State $N_s \gets i$
                \State $v_\mathrm{end} \gets \begin{cases}
                                                 0, & \text{\small if $s_{N_s}$ is terminal}\\
                                                 V(s_{N_s},\theta), & \text{\small otherwise}
                                              \end{cases}$
                \For{$i \in 0:N_s-1$}
                    \State $z_i \gets \sum_{k=i}^{N_s-1}\gamma^{k-i}r_k + \gamma^{N_s-i}v_\mathrm{end}$ 
					\State \Call{AddSampleToMemory}{$s_i,\pi_i,z_i$}
                \EndFor
            \EndWhile
        \EndFunction
    \end{algorithmic}
\end{algorithm}


\section{Implementation}
\label{sec:implementation}

\begin{figure*}[!t]
    \centering
        \subfloat[Continuous highway driving case.]{\includegraphics[width=1.99\columnwidth]{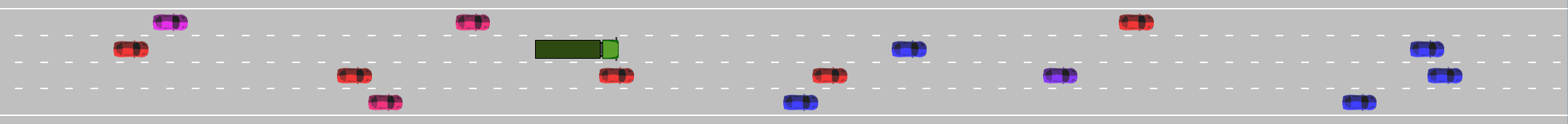}%
        \label{fig:contScenario}}
        \vspace{-7pt} 
        \captionsetup[subfloat]{captionskip=-13pt}
        \subfloat[Highway exit case.]{\includegraphics[width=1.99\columnwidth]{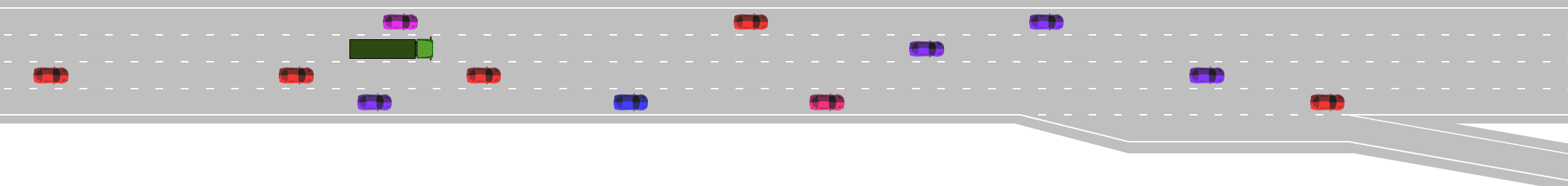}%
        \label{fig:exitScenario}}
    \vspace{-10pt}
    \caption{Examples of the two test cases. (a) shows an initial state for the continuous highway driving case, whereas (b) shows the exit case, when the ego vehicle is approaching the exit on the right side of the road. The ego vehicle is the green truck, whereas the color of the surrounding vehicles represent the aggressiveness level of their corresponding driver models, see Sect.~\ref{sec:initialization}. Red is an aggressive driver, blue is a timid driver, and the different shades of purple represent levels in between.}
    \label{fig:testScenarios}
\end{figure*}

The presented framework for combining planning and reinforcement learning can be applied to autonomous driving. In this study, the properties of the framework were investigated for two highway driving cases, which are illustrated in Fig.~\ref{fig:testScenarios}. The first case involves navigating in traffic as efficiently as possible, whereas the second case involves exiting on an off-ramp. This section starts with describing the driver and physical modeling of the the cases, which is used both as a generative model and for simulating the environment, and is then followed by a description of how the proposed framework was applied, how the simulations were set up, and how the baseline methods were implemented.
The design of the test cases was inspired by Sunberg et al.~\cite{Sunberg2017}.

\subsection{Driver Modeling}
\label{sec:driverModeling}
The Intelligent Driver Model (IDM) was used to govern the longitudinal motion of each vehicle~\cite{IDM}. The longitudinal acceleration $\dot{v}_\mathrm{IDM}$ is determined by
\begin{align}
\label{eq:IDM}
	\dot{v}_\mathrm{IDM} = a \bigg( 1 - \left( \frac{v_x}{v_\mathrm{set}} \right)^4 - \left( \frac{d^*(v_x,\Delta v_x)}{d} \right)^2 \bigg),
\end{align}
where $v_x$ is the longitudinal speed of the vehicle, and $d^*$ is the desired distance to the vehicle ahead, given by
\begin{align}
\label{eq:IDM_2}
		d^*(v_x,\Delta v_x) = d_0 + v_xT_\mathrm{set} + v_x\Delta v_x / (2 \sqrt{ab}).
\end{align}
The acceleration is a function of the vehicle speed $v_x$, the distance to the vehicle ahead $d$, and the speed difference (approach rate) $\Delta v_x$.
The parameters of the model are the set speed $v_\mathrm{set}$, the set time gap $T_\mathrm{set}$, the minimum distance $d_0$, the maximum acceleration $a$, and the desired deceleration $b$.

Noise was added to the acceleration of the vehicles $\dot{v}_x$, by setting
\begin{align}
    \dot{v}_x = \dot{v}_\mathrm{IDM} + \frac{\sigma_\mathrm{vel}}{\Delta t}w,
\end{align}
where $w$ is a normally distributed random variable with unit standard deviation and zero mean, which is independent for each vehicle. The parameter $\sigma_\mathrm{vel}$ is the standard deviation of the velocity noise and $\Delta t$ is the time step of the simulation. No noise was added to the ego vehicle acceleration.

The Minimizing Overall Braking Induced by Lane changes (MOBIL) strategy was used to model the lane changes of the surrounding vehicles~\cite{MOBIL}. This model makes lane changing decisions with the goal of maximizing the acceleration of all the involved traffic participants at every time step. A lane change is only allowed if the induced acceleration of the following vehicle in the new lane $a_\mathrm{n}$ fulfills a safety criterion, $a_\mathrm{n} > -b_\mathrm{safe}$. The IDM is used to predict the accelerations of the neighboring vehicles. Then, a lane change is performed if
\begin{align}
	\tilde{a}_\mathrm{e} - a_\mathrm{e} + p \left( (\tilde{a}_\mathrm{n} - a_\mathrm{n}) + (\tilde{a}_\mathrm{o} - a_\mathrm{o}) \right) > a_\mathrm{th},
\end{align}
where $a_\mathrm{e}$, $a_\mathrm{n}$, and $a_\mathrm{o}$ are the accelerations of the ego vehicle, the following vehicle in the target lane, and the following vehicle in the current lane, respectively, if no lane change is performed. Moreover, $\tilde{a}_\mathrm{e}$, $\tilde{a}_\mathrm{n}$, and $\tilde{a}_\mathrm{o}$ are the corresponding accelerations if the ego vehicle changes lane. A politeness factor $p$ controls how much the gains and losses of the surrounding vehicles are valued. The lane change is done if the sum of the weighted acceleration changes is higher than a threshold $a_\mathrm{th}$.
Finally, if lanes are available both to the left and to the right, the same criterion is applied to both options. If these are both fulfilled, the model chooses the option with the highest acceleration gain.

\subsection{Physical Modeling}
\label{sec:physicalModeling}

Both test cases took place on a straight one-way highway with four lanes. The longitudinal dynamics assumed a constant acceleration, which means that the longitudinal position and speed of a vehicle, $x$ and $v_x$, were updated according to
\begin{align}
    x' & = x + v_x\Delta t + \frac{1}{2}\dot{v}_x\Delta t^2,\\
    v_x' & = v_x + \dot{v}_x\Delta t.
\end{align}
Furthermore, the braking acceleration was limited to $b_\mathrm{max}$.

The lateral dynamics assumed a constant lateral speed $v_y$, which means that the lateral position of a vehicle $y$, was updated according to
\begin{align}
    y' = y + v_y\Delta t.
\end{align}
When a lane change was requested, $v_y$ was set to $\pm v_y^\mathrm{lc}$, where the sign depends on the intended lane change direction. Otherwise, it was set to 0. Table~\ref{tab:simParams} provides the parameter values.

\subsection{POMDP Formulation}
\label{sec:pomdpFormulation}

This section describes how the decision making problem for the two highway driving cases was formulated as a POMDP, how the state of the system was estimated from observations, and how Algorithm~\ref{alg:treeSearch} was used to make the decisions. Table~\ref{tab:simParams} provides the parameter values.

\subsubsection{State space, $\mathcal{S}$}
The state of the system,
\begin{align}
    s = (s_\mathrm{term},\{(s^p_i,s_i^d)\}_{i\in0,\dots,N_\mathrm{veh}}),
\end{align}
consists of the physical state $s_i^p$ and the driver state (driver model parameters) $s_i^d$ of the ego vehicle, and the $N_\mathrm{veh}$ surrounding vehicles in a traffic scene. There is also a Boolean state $s_\mathrm{term}$, which takes the value 1 when a terminal state is reached, and otherwise 0.
The physical state consists of the longitudinal and lateral position, and speed, of each vehicle,
\begin{align}
    s^p_i = (x_i,y_i,v_{x,i},v_{y,i}).
\end{align}
The driver state is described by the driver model parameters,
\begin{align}
    s^d_i = (v_{\mathrm{set},i}, T_{\mathrm{set},i}, d_{0,i}, a_i, b_i, p_i, a_{\mathrm{th},i}, b_{\mathrm{safe},i}).
\end{align}

\subsubsection{Action space, $\mathcal{A}$}

Since this study concerns tactical driving decisions, a high level action space is used. At every time step, the agent can choose between $m_\mathrm{act}=5$ different actions; keep its current driver state $a_1$, decrease or increase the setpoint of the adaptive cruise controller (ACC) $a_2$, $a_3$, and change lanes to the right or to the left $a_4$, $a_5$. Table~\ref{tab:actions} provides a summary of the available actions.
The decision is then forwarded to a lower level operational decision making layer, which handles the detailed execution of the requested maneuver.

In this study, a simplified operational decision making layer is used, where the ACC consists of the IDM.
In short, increasing the ACC setpoint corresponds to increasing the requested speed or decreasing the time gap, whereas decreasing the ACC setpoint has the opposite effect.
More specifically, when $a_3$ is chosen and the set speed of the IDM  $v_\mathrm{set}$ is less than the speed desired by a higher level strategic decision making layer $v_\mathrm{des}$, then $v_\mathrm{set}$ is increased by $\Delta v_\mathrm{set}$. However, if $v_\mathrm{set}=v_\mathrm{des}$, then the set time gap of the IDM $T_\mathrm{set}$ is decreased with $\Delta T_\mathrm{set}$. Similarly, if $a_2$ is chosen and $T_\mathrm{set} < T_\mathrm{max}$, where $T_\mathrm{max}$ is the maximum allowed time gap of the ACC, then $T_\mathrm{set}$ is increased by $\Delta T_\mathrm{set}$. However, if $T_\mathrm{set} = T_\mathrm{max}$, then $v_\mathrm{set}$ is decreased by $\Delta v_\mathrm{set}$. When action $a_4$ or $a_5$ are chosen, the vehicle either starts a lane change, continues a lane change or aborts a lane change, i.e., moves to the right or the left respectively by setting $v_y = \pm v_y^\mathrm{lc}$. When a lane change is performed, the set speed is reset to $v_\mathrm{set}=v_\mathrm{des}$ and the set time gap $T_\mathrm{set}$ is set to the actual time gap. Decisions were taken at an interval of $\Delta t$.

At every time step, the action space is pruned, in order to remove actions that lead to collisions. A lane change action is only allowed if the ego vehicle or the new trailing vehicle need to brake with an acceleration lower than $a_\mathrm{min}$ to avoid a collision. Since the IDM itself is also crash-free, the ego vehicle will never cause a collision.
Furthermore, a minimum time gap $T_\mathrm{min}$ setpoint of the ACC is used. Therefore, if $T_\mathrm{set} = T_\mathrm{min}$, then $a_3$ is not considered. Moreover, if a lane change is ongoing, i.e., the vehicle is positioned between two lanes, only actions $a_4$ and $a_5$ are considered, i.e., to continue the lane change or change back to the previous lane.

\begin{table}[!t]
	\caption{Action space of the agent.}
	\label{tab:actions}
	\centering
	\begin{tabular}{ll}
	\toprule
		$a_1$ & Stay in current lane, keep current ACC setpoint\\
	    $a_2$ & Stay in current lane, decrease ACC setpoint\\
		$a_3$ & Stay in current lane, increase ACC setpoint\\
		$a_4$ & Change lanes to the right, keep current ACC setpoint\\
		$a_5$ & Change lanes to the left, keep current ACC setpoint\\
		\bottomrule
	\end{tabular}
\end{table}

\subsubsection{Reward model, $R$}
The objective of the agent is to navigate the highway safely and efficiently. Since safety is handled by a crash-free action set, a simple reward function that tries to minimize the time and the number of lane changes is used. Normally, at every time step, a positive reward of $1-|\frac{v_\mathrm{ego}-v_\mathrm{des}}{v_\mathrm{des}}|$ is given, which penalizes deviations from the desired speed. If a lane change is initiated, a negative reward of $c_\mathrm{lc}$ is added. Finally, for the case with the highway exit, a reward $r_\mathrm{term}$ is added when the agent transitions to a terminal state. If the exit is reached at this time, then $r_\mathrm{term}=\gamma\frac{1}{1-\gamma}$, which (from a geometric sum) corresponds to that the vehicle would have continued to drive forever and gotten the reward $1$ at every subsequent step. If the exit is not reached, then $r_\mathrm{term}=0$.

\subsubsection{State transition model, $T$}
The state transition model is implicitly defined by a generative model.

\subsubsection{Generative model, $G$}
The combination of the IDM and MOBIL model, and the physical model are used as a generative model $G$, where $s' \sim G(s,a)$.
The same generative model is used in the tree search of Algorithm~\ref{alg:treeSearch}. 

\subsubsection{Observation space, $\mathcal{O}$}
The observations $o$ consists of the physical states of the surrounding vehicles, and the physical and driver state of the ego vehicle, i.e., $o = (s_0^p,s_0^d,\{s_i^p\}_{i\in1,\dots,N_\mathrm{veh}})$. The driver states of the surrounding vehicles are not included in the observation.

\subsubsection{Observation model, $O$}
A simplified sensor model was used in this study. The physical state of all vehicles that are positioned closer than $x_\mathrm{sensor}$ of the ego vehicle is assumed to be observed exactly, whereas vehicles further away are not detected at all.

\subsubsection{Belief state estimator}
\label{sec:behaviorEstimation}

The driver state of the surrounding vehicles cannot be directly observed, but it can be estimated from the vehicles' physical state history by using a particle filter~\cite{Sunberg2017}. A particle $\hat{s}^d$ represents the value of the driver model parameters of the observed surrounding vehicles. At a given time step, a collection of $M$ particles $\{\hat{s}^d_k\}_{k=1}^M$ and their associated weights $\{W_k\}_{k=1}^M$ describe the belief of the driver model parameters. At the next time step, after action $a$ has been taken, the belief is updated by sampling $M$ new particles with a probability that is proportional to the weights. Then, new states are generated by $\hat{s}_k' = G((s^p,s_0^d,\hat{s}_k^d),a)$. Note that there is a component of noise in $G$, see Sect.~\ref{sec:driverModeling}. The new weights are calculated from the new observation, as the approximate conditional probability
\begin{align}
    W_k' = 
    \left\{
    \begin{array}{@{}ll@{}}
        \exp{\left(-\frac{(v'-\hat{v}')^2}{2\sigma_\mathrm{vel}^2}\right)} & \text{if } y'=\hat{y}'\\
        \gamma_\mathrm{lane} \exp{\left( -\frac{(v'-\hat{v}')^2}{2\sigma_\mathrm{vel}^2} \right)} & \text{otherwise}
   \end{array}
   \right\}
    \appropto
    \mathrm{Pr}\left( \hat{s}_k \mid o \right), \nonumber \\[-10 pt]
    & \label{eq:W}
\end{align}
where $v'$ and $y'$ are given by the observation, $\hat{v}'$ is taken from $\hat{s}_k'$, and $\gamma_\mathrm{lane}$ is a parameter that penalizes false lane changes. Furthermore, Gaussian noise with a standard deviation that is proportional to the sample standard deviation of the current $M$ particles is added to 10\% of the new samples, in order to prevent particle deprivation.

In Algorithm~\ref{alg:treeSearch}, the estimated most likely state is used as input. The function is called with $\Call{SelectAction}{s_0,n,\theta}$, where $s_0 = (s_\mathrm{term},s^p,s_0^d,\hat{s}^d_\mathrm{max})$ consists of the terminal state, the observed physical state, the observed ego driver state, and the particle with the highest weight. This particle represents the most likely driver model state, given by $\hat{s}^d_\mathrm{max} = \hat{s}^d_{\argmax_k W_k}$.

\subsection{Neural Network Architecture and Training Process}
\label{sec:nnArchitectureAndTraining}
A neural network estimates the prior probabilities of taking different actions and the value of the current state. 
In this implementation, before the state $s$ is passed through the neural network, it is converted to $\xi$, where all states are normalized, i.e., $\xi_* \in [\minus 1,1]$, and the positions and velocities of the surrounding vehicles are expressed relative to the ego vehicle.
There are $m_\mathrm{ego}=7$ inputs that describe the ego vehicle state and $m_\mathrm{veh}=4$ inputs that describe the state of each surrounding vehicle.
The first elements, $\xi_1$ to $\xi_7$, describe the state of the ego vehicle, whereas $\xi_{7i+1}$, $\xi_{7i+2}$, $\xi_{7i+3}$, and $\xi_{7i+4}$, for $i=1, \dots, N_\mathrm{max}$, represent the relative state of the surrounding vehicles. If there are less than $N_\mathrm{max}$ vehicles in the sensing range, the remaining inputs are padded with dummy values, $\xi_{7i+1}=\minus1$ and $\xi_{7i+2}=\xi_{7i+3}=\xi_{7i+4}=0$, which will not affect the output of the network, see below.
Table~\ref{tab:nnState} describes how $\xi$ is calculated and the values of the normalization parameters are given in Table~\ref{tab:simParams}.

In a previous study~\cite{Paper2}, we introduced a temporal convolutional neural network (CNN) architecture, which simplifies and speeds up the training process by applying CNN layers to the part of the input that consists of interchangeable objects, in this case surrounding vehicles. 
The input that describes the surrounding vehicles is passed through CNN layers, with a design that results in identical weights for each vehicle, and finally a max pooling layer creates a translational invariance between the vehicles.
With this structure, the output will not depend on how the vehicles are ordered in the input vector. It also removes the problem with a fix input vector size, since it can simply be made larger than necessary and padded with dummy values for the extra vehicle slots. The extra input will then be removed by the max pooling layer.

\begin{figure*}[!t]
    \centering
        \includegraphics[width=1.99\columnwidth]{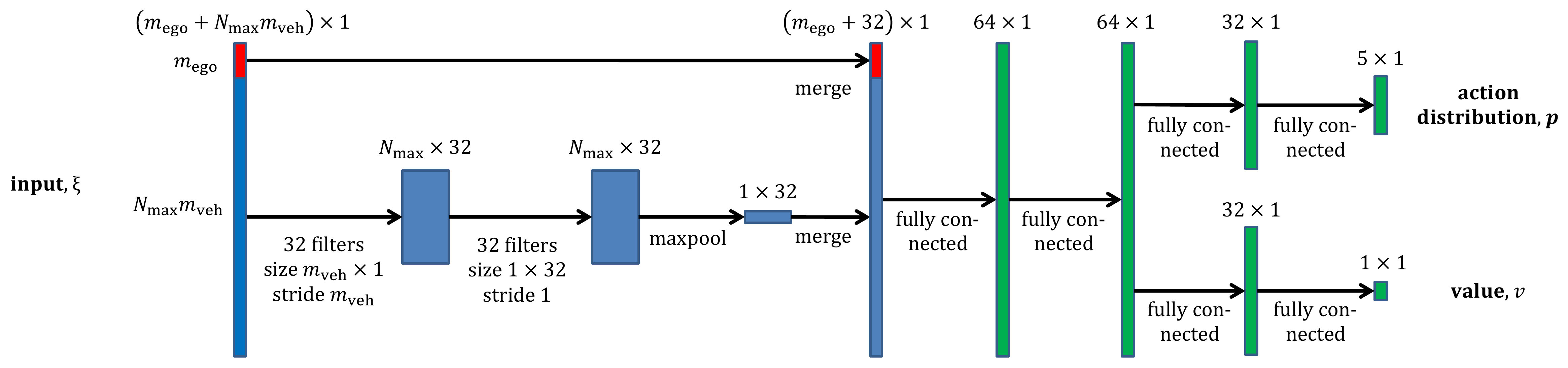}
        \caption{The figure illustrates the neural network architecture that was used in this study. The convolutional and max pooling layers create a translational invariance between the input from different surrounding vehicles, which makes the ordering and the number of vehicles irrelevant.}
    \label{fig:neuralNetworkArchitecture}
\end{figure*}

The neural network architecture, shown in Fig.~\ref{fig:neuralNetworkArchitecture}, was used in this study and includes the described CNN architecture.
In short, the input that describe the state of the surrounding vehicles is passed through two convolutional layers, followed by a max pooling layer. This is then concatenated with the input that describes the ego vehicle state and passed through two fully connected layers, before being split up into two parallel heads, which estimate the action distribution $\bm{p}(s,\theta)$ and the value $V(s,\theta)$ of the input state $s$.
All layers have ReLU activation functions, except for the action head, which has a softmax activation function, and the value head, which has a sigmoid activation function. Finally, the value output is scaled with the factor $1/(1-\gamma)$ (since this is the maximum possible value of a state).

\begin{table}[!t]
	\renewcommand{\arraystretch}{1.1}
	\caption{Input to the neural network $\xi$.
	}
	\label{tab:nnState}
	\centering
	\begin{tabular}{ll}
		\toprule
	    Ego lane & $\xi_1 = 2y_0 / y_\mathrm{max} - 1$\\
	    Ego vehicle speed & $\xi_2 = 2v_{x,0} / v_{x,0}^\mathrm{max} - 1$\\
	    Lane change state & $\xi_3 = \sgn{(v_{y,0})}$\\
	    Ego vehicle set speed & $\xi_4 = 2v_{\mathrm{set},0} / v_{x,0}^\mathrm{max} - 1$\\
	    Ego vehicle set time gap & $\xi_5 = \frac{T_{\mathrm{set},0} - (T_\mathrm{max}+T_\mathrm{min})/2} {(T_\mathrm{max} - T_\mathrm{min})/2}$\\
	    Distance to exit & $\xi_6 = 1 - 2x_0 / x_\mathrm{exit}$\\ 
	    Terminal state & $\xi_7 = s_\mathrm{term}$ \\
		Relative long. position of vehicle $i$ & $\xi_{7i+1} = (x_i-x_0) / x_\mathrm{sensor}$\\
		Relative lat. position of vehicle $i$ & $\xi_{7i+2} = (y_i-y_0) / y_\mathrm{max}$\\
		Relative speed of vehicle $i$ & $\xi_{7i+3} = \frac{v_{x,i} - v_{x,0}}{v_\mathrm{set}^\mathrm{max} - v_\mathrm{set}^\mathrm{min}}$\\ 
		Lane change state of vehicle $i$ & $\xi_{7i+4} = \sgn{(v_{y,i})}$\\
		\bottomrule
	\end{tabular}
\end{table}

Algorithm~\ref{alg:training} was used to collect training samples. When an episode was finished, the $n_\mathrm{new}$ samples were added to a replay memory of size $M_\mathrm{replay}$. Learning started after $N_\mathrm{start}$ samples had been added. Then, after each finished episode, the network was trained on $n_\mathrm{new}$ mini-bathes, with size $M_\mathrm{mini}$, uniformly drawn from the memory. The loss was calculated by using Eq.~\ref{eq:loss} and the neural network parameters $\theta$ were optimized by stochastic gradient descent, with learning rate $\eta$ and momentum $\mu$. The parameters of Algorithm~\ref{alg:treeSearch} and the training process are shown in Table~\ref{tab:mctsAndNnParameters}.
In order to speed up the training process, the algorithm was parallelized. Twenty workers simultaneously ran simulations to generate training data. They all shared the same neural network and update process. Worker calls to $f_\theta(s)$ were queued and passed to the neural network in batches.

\begin{table}[!t]
	\caption{Parameters of the MCTS and the neural network training.}
	\label{tab:mctsAndNnParameters}
	\centering
	\begin{tabular}{llr}
		\toprule
		MCTS iterations & $n$ & $2{,}000$\\
		Exploration parameter & $c_\mathrm{puct}$ & $0.1$\\
		Progressive widening linear param. & $k$ & $1.0$\\
		Progressive widening exponent param. & $\alpha$ & $0.3$\\
		Temperature parameter, & $\tau$ & $1.1$\\
		Dirichlet noise parameter & $\beta$ & $1.0$\\
		Noise proportion parameter & $\epsilon$ & $0.25$\\
		
		\midrule
		
		Training start & $N_\mathrm{start}$ & $20{,}000$\\
		Replay memory size & $M_\mathrm{replay}$ & $100{,}000$\\
		Mini-batch size & $M_\mathrm{mini}$ & $32$\\
		
		Loss function weights & $(c_1,c_2,c_3)$ & $(100, 1, 0.0001)$\\

		Learning rate & $\eta$ & $0.01$\\
		Momentum & $\mu$ & $0.9$\\
		
		\bottomrule
	\end{tabular}
\end{table}

\begin{table}[!t]
	\caption{Various simulation parameters.}
	\label{tab:simParams}
	\centering
	\begin{tabular}{llr}
		\toprule
		Velocity noise standard deviation & $\sigma_\mathrm{vel}$ & $0.5$ m/s\\
		Physical braking limit & $b_\mathrm{max}$ & $8.0$ m/s\textsuperscript{2}\\
		Simulation time step & $\Delta t$ & $0.75$ s\\
		Lateral lane change speed & $v_y^\mathrm{lc}$ & $0.67$ lanes/s\\
		
		Minimum set time gap & $T_\mathrm{min}$ & $0.5$ s\\
		Maximum set time gap & $T_\mathrm{max}$ & $2.5$ s\\
		Step set time gap & $\Delta T_\mathrm{set}$ & $1.0$ s\\
		Step set speed & $\Delta v_\mathrm{set}$ & $2.0$ m/s\\
		Desired speed of strategic layer & $v_\mathrm{des}$ & $25$ m/s\\
		
		Lane change penalty & $c_\mathrm{lc}$ & $\minus 0.03$\\
		
		Discount factor & $\gamma$ & $0.95$\\ 
		
		Surrounding sensor range & $x_\mathrm{sensor}$ & $100$ m\\
		
		Maximum number of vehicles & $N_\mathrm{max}$ & $20$\\
		Exit lane position & $x_\mathrm{exit}$ & $1{,}000$ m\\
		Initial ego speed & $v_{x,0}$ & $20$ m/s\\
		
		Number of particles & $M$ & $500$\\
		Particle filter lane factor & $\gamma_\mathrm{lane}$ & $0.2$\\
		
		State normalization lane & $y_\mathrm{max}$ & $4$ lanes\\
		Maximum ego vehicle speed & $v_{x,0}^\mathrm{max}$ & $25$ m/s\\
		Minimum set speed (timid driver) & $v_\mathrm{set}^\mathrm{min}$ & $19.4$ m/s\\
		Maximum set speed (aggressive driver) & $v_\mathrm{set}^\mathrm{max}$ & $30.6$ m/s\\

		\bottomrule
	\end{tabular}
\end{table}

\subsection{Episode Implementation}
\label{sec:initialization}

As mentioned above, both the test cases consisted of a straight one-way highway with four lanes. Overtaking was allowed both on the left and the right side of another vehicle. The continuous driving case ended after $200$ time steps (with $s_\mathrm{term}=0)$, whereas the exit case ended when the ego vehicle reached the exit position longitudinally, i.e., $x_0 \geq x_\mathrm{exit}$ (with $s_\mathrm{term}=1$). To successfully reach this exit, the ego vehicle had to be in the rightmost lane at this point.
The ego vehicle, a $12.0$ m long truck, started in a random lane for the continuous case and in the leftmost lane for the exit case, with an initial velocity of $v_{x,0}$. The allowed speed limit for trucks was $25$ m/s, hence $v_\mathrm{des} = 25$ m/s. In short, around $20$ passenger cars, with a length of $4.8$ m, were placed on the road, where slower vehicles were positioned in front of the ego vehicle and faster vehicles were positioned behind. An example of an initial state is shown in Fig.~\ref{fig:contScenario}. The details on how the initial states were created are described below.

The surrounding vehicles were controlled by the IDM and MOBIL model. The marginal distribution of the model parameters were uniformly distributed between aggressive and timid driver parameters, which were slightly adapted from Kesting et al.~\cite{ModelParameters} and shown in Table~\ref{tab:idmMobilParameters}. The main difference is that the politeness factor is here significantly reduced, to create a more challenging task, where slow drivers do not always try to move out of the way.
In order to create a new driver model, values were drawn from a Gaussian copula, which had a covariance matrix with 1 along the diagonal and a correlation of $\rho = 0.75$ elsewhere. These values were then scaled and translated to the range between aggressive and timid driver parameters.

In order to create the initial state of the simulation, at first only the ego vehicle was placed on the road and the simulation was run for $n_\mathrm{init}=200$ steps. During this phase, the ego vehicle was controlled by the IDM and it made no lane changes. At every time step, a new vehicle with random parameters was generated. If it was faster than the ego vehicle, it was inserted $300$ m behind it, and if it was slower, $300$ m in front of it. Furthermore, it was placed in the lane that had the largest clearance to any other vehicle. However, if $d^*$ (see Eq.~\ref{eq:IDM_2}) of the new vehicle, or the vehicle behind it, was less than the actual gap, the vehicle was not inserted. At most $N_\mathrm{max}$ vehicle were added. The state after these $n_\mathrm{init}$ steps was then used as the initial state of the new episode. The ego vehicle driver state was initialized to the same parameters as a `normal' driver (Table~\ref{tab:idmMobilParameters}).

\begin{table}[!t]
	\caption{IDM and MOBIL model parameters for different driver types.}
	\label{tab:idmMobilParameters}
	\centering
	\begin{tabular}{llrrr}
		\toprule
		& & Normal & Timid & Aggressive\\
		\hline
		Desired speed (m/s) & $v_\mathrm{set}$ & $25.0$ & $19.4$ & $30.6$ \\
		Desired time gap (s) & $T_\mathrm{set}$ & $1.5$ & $2.0$ & $1.0$\\
		Minimum gap distance (m) & $d_0$ & $2.0$ & $4.0$ & $0.0$\\
		Maximal acceleration (m/s\textsuperscript{2}) & $a$ & $1.4$ & $0.8$ & $2.0$\\
		Desired deceleration (m/s\textsuperscript{2}) & $b$ & $2.0$ & $1.0$ & $3.0$\\
		\midrule
		Politeness factor & $p$ & $0.05$ & $0.1$ & $0.0$\\
		Changing threshold (m/s\textsuperscript{2}) & $a_\mathrm{th}$ & $0.1$ & $0.2$ & $0.0$\\
		Safe braking (m/s\textsuperscript{2}) & $b_\mathrm{safe}$ & $2.0$ & $1.0$ & $3.0$\\
		\bottomrule
	\end{tabular}
\end{table}

\subsection{Baseline Methods}
\label{sec:baselines}

The performance of the framework that is introduced in this paper is compared to two baseline methods. For both test cases, standard MCTS with progressive widening, described in Sect.~\ref{sec:mcts}, is used as a baseline, with the same POMDP formulation that is described in Sect.~\ref{sec:pomdpFormulation}. Furthermore, for a fair comparison, the same parameter values as for the introduced framework is used, described in Table~\ref{tab:mctsAndNnParameters}, and the exploration parameter is set to $c_\mathrm{uct} = c_\mathrm{puct}$. Rollouts are done using the IDM and MOBIL model, with the `normal' driver parameters of Table~\ref{tab:idmMobilParameters}, and a rollout depth of 20 time steps. After $n$ iterations are performed, the action with the highest visit count is chosen.

For the continuous highway driving case, a second baseline is the IDM and MOBIL model, with the `normal' driver parameters. A similar model is used as a second baseline for the highway case with an exit. Then, the driver follows the IDM longitudinally and changes lanes to the right as soon as the safety criterion of the MOBIL model is fulfilled.


\section{Results}
\label{sec:results}

The results show that the agents that were obtained by applying the proposed framework to the continuous highway driving case and the highway exit case outperformed the baseline methods.
This section presents further results for the two test cases, together with an explanation and brief discussion on some of the characteristics of the results.
A more general discussion follows in Sect.~\ref{sec:discussion}.
The decision making agent that was created from the presented framework is henceforth called the MCTS/NN agent (where NN refers to neural network), whereas the baseline methods are called the standard MCTS agent and the IDM/MOBIL agent.

For both test cases, the MCTS/NN agent was trained in a simulated environment (Sect.~\ref{sec:nnArchitectureAndTraining}). At every $20{,}000$ added training samples, an evaluation phase was run, where the agent was tested on $100$ different episodes. These evaluation episodes were randomly generated (Sect.~\ref{sec:initialization}), but they were identical for all the evaluation phases and for the different agents.

\subsection{Continuous Highway Driving Case}
\label{sec:resultsContinuous}

\begin{figure}[!t]
    \centering
        \includegraphics[width=0.99\columnwidth]{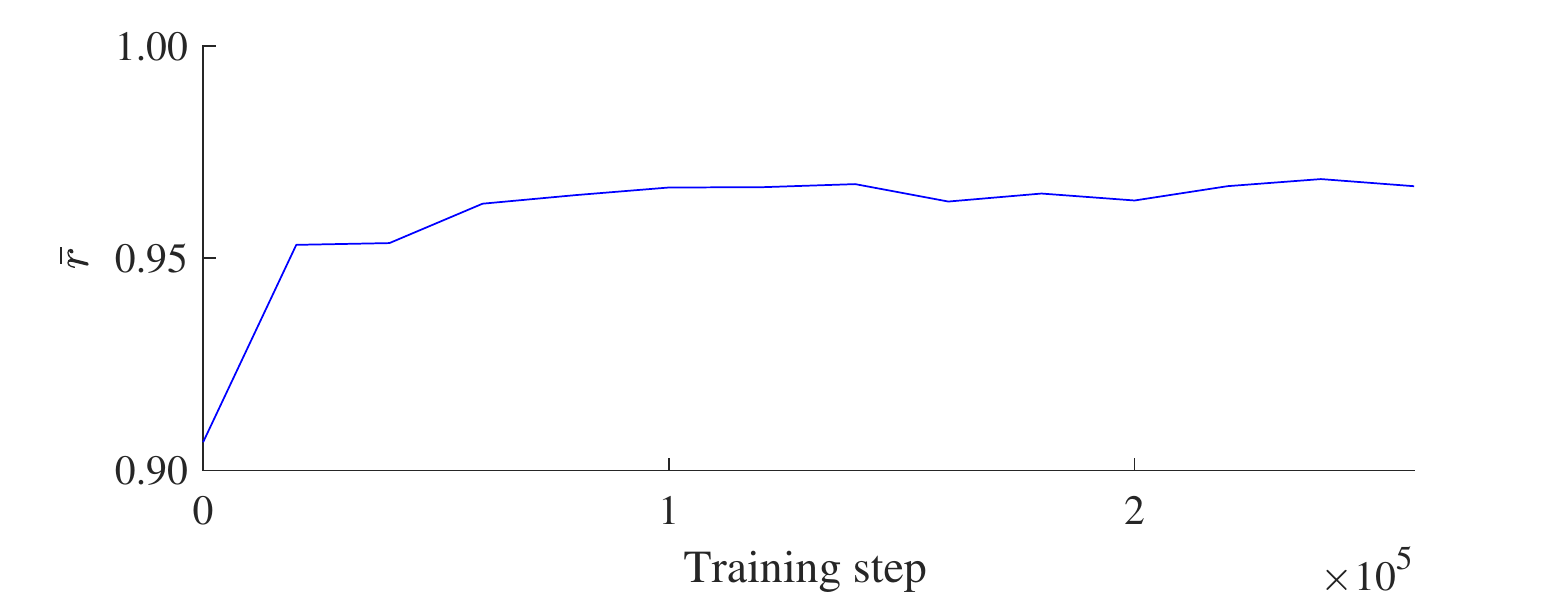}
        \caption{Average reward per step $\bar{r}$ during the evaluation episodes, as a function of the number of training steps, for the continuous highway driving case.}
    \label{fig:contRewardEvolution}
\end{figure}

Fig.~\ref{fig:contRewardEvolution} shows the average reward $\bar{r}$ per step that was received during the evaluation episodes, as a function of the number of added training samples, henceforth called training steps. The maximum possible reward for every step is 1, and it is decreased when the agent deviates from the desired speed and/or make lane changes. The agent performed relatively well even before the training had started, due to its planning component. However, with only $20{,}000$ training steps, the agent learned to perform significantly better. The average received reward then increased slightly with more training, until around $100{,}000$ steps, where the performance settled.

\begin{figure}[!t]
    \centering
		\includegraphics[width=\columnwidth]{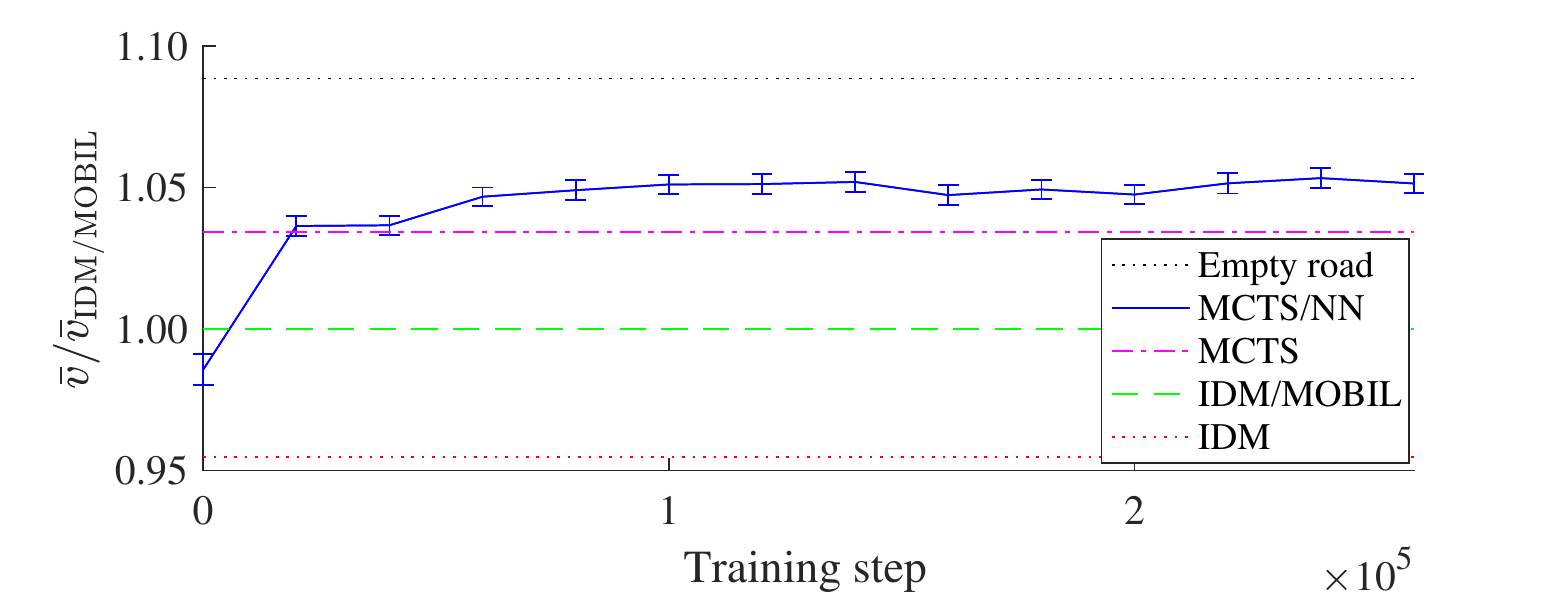}
	\caption{Mean speed $\bar{v}$ during the evaluation episodes for the continuous highway driving case, normalized with the mean speed of the IDM/MOBIL model $\bar{v}_\mathrm{IDM/MOBIL}$.
	The error bars indicate the standard error of the mean for the MCTS/NN agent, i.e., $\sigma_\mathrm{sample}/\sqrt{100}$, where $\sigma_\mathrm{sample}$ is the standard deviation of the $100$ evaluation episodes.}
	\label{fig:contMeanSpeedEvolution}
\end{figure}

A comparison of the average speed $\bar{v}$ and the action distribution during the evaluation episodes was made for the MCTS/NN agent and the baseline methods. Fig.~\ref{fig:contMeanSpeedEvolution} shows how $\bar{v}$ varies during the training of the agent, normalized with the mean speed of the IDM/MOBIL model $\bar{v}_\mathrm{IDM/MOBIL}$. For reference, the figure displays the average speed when applying the IDM, which always stays in its original lane during the whole episode, and can therefore be considered as a minimum performance. 
The ideal mean speed ($25$ m/s) for when the road is empty is also indicated, which is naturally not achievable in these episodes due to the surrounding traffic.
The figure shows that the standard MCTS agent outperformed the IDM/MOBIL agent, and that the MCTS/NN agent quickly learned to match the performance of the MCTS agent and then surpassed it after around $60{,}000$ training steps. Table~\ref{tab:contActionDistribution} shows the action distribution for the baseline methods and for the MCTS/NN agent after $250{,}000$ training steps. The trained MCTS/NN agent and the IDM/MOBIL agent performed lane changes at about the same rate, whereas the standard MCTS agent made significantly more lane changes. It also changed its ACC state more than the trained MCTS/NN agent.

\begin{table}[!t]
	\caption{Action distribution for the continuous highway driving case.}
	\label{tab:contActionDistribution}
	\centering
	\begin{tabular}{lccccc}
		\toprule
		& Idle & ACC down & ACC up & Right & Left\\
		IDM/MOBIL & $95.2\%$ & - & - & $2.3\%$ & $2.5\%$\\
		MCTS & $61.8\%$ & $12.4\%$ & $13.8\%$ & $6.0\%$ & $6.0\%$\\
		MCTS/NN & $90.4\%$ & $1.1\%$ & $3.9\%$ & $2.1\%$ & $2.5\%$\\
		\bottomrule
	\end{tabular}
\end{table}

\begin{figure}[!t]
    \centering
        \subfloat[$t=0$]{\includegraphics[width=0.45\columnwidth]{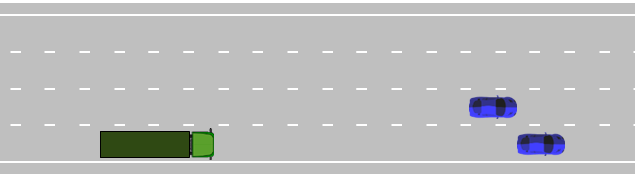}%
        \label{fig:doubleLaneChangeStart}}
        \hfil
        \subfloat[$t=15$ s, IDM/MOBIL]{\includegraphics[width=0.45\columnwidth]{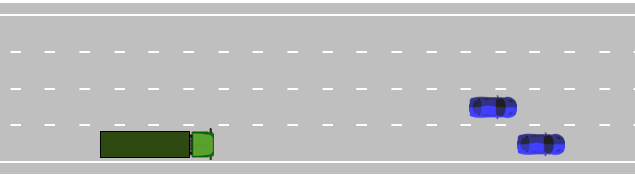}%
        \label{fig:doubleLaneChangeRef}}
        \hfil
        \subfloat[$t=15$ s, MCTS]{\includegraphics[width=0.45\columnwidth]{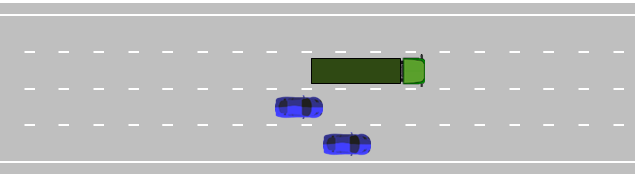}%
        \label{fig:doubleLaneChangeStartMCTS}}
        \hfil
        \subfloat[$t=15$ s, MCTS/NN]{\includegraphics[width=0.45\columnwidth]{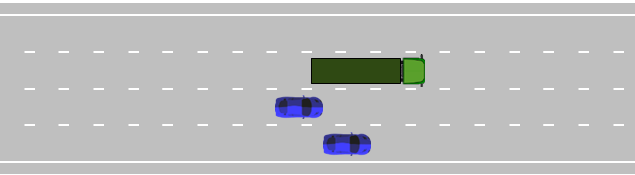}%
        \label{fig:doubleLaneChangeStartMCTSNN}}
    \caption{Example of a situation where planning is necessary.
    The initial state is shown in (a) and the state after $15$ s is shown for the three agents in (b), (c), and (d). The green truck is the ego vehicle.}
    \label{fig:doubleLaneChange}
\end{figure}

A key difference between the compared methods is highlighted in Fig.~\ref{fig:doubleLaneChange}, which shows a situation where planning is required. The ego vehicle is here placed behind two slow vehicles in adjacent lanes, with timid driver parameters. The best behavior for the ego vehicle is to make two lane changes to the left, in order to overtake the slow vehicles. Both the standard MCTS agent and the trained MCTS/NN agents solved this situation, whereas the IDM/MOBIL agent got stuck in the original lane.

\begin{figure}[!t]
    \centering
        \includegraphics[width=0.99\columnwidth]{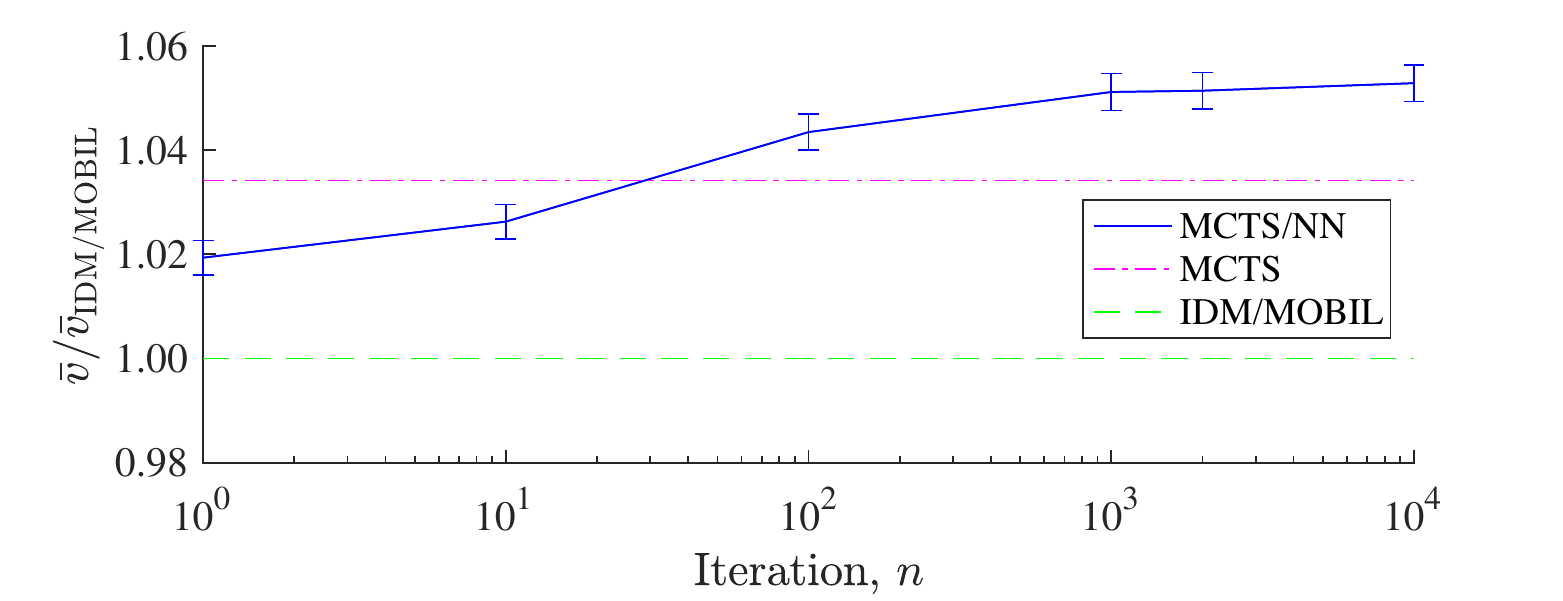}
        \caption{Normalized mean speed as a function of the number of MCTS iterations $n$ for the trained MCTS/NN agent, in the continuous highway driving case. The error bars indicate the standard error of the mean.}
    \label{fig:contPerformanceSearches}
\end{figure}

Finally, the effect of the number of MCTS iterations $n$ on the performance of the trained agent was investigated, which is shown in Fig.~\ref{fig:contPerformanceSearches}. To execute just one iteration corresponds to using the policy that was learned by the neural network, which performed better than the IDM/MOBIL agent. At around $30$ iterations the MCTS/NN agent surpassed the standard MCTS agent, which used $2{,}000$ iterations, and at $n=1{,}000$ the performance settled.

\subsection{Highway Exit Case}
\label{sec:resultsExit}

\begin{figure}[!t]
    \centering
        \includegraphics[width=0.99\columnwidth]{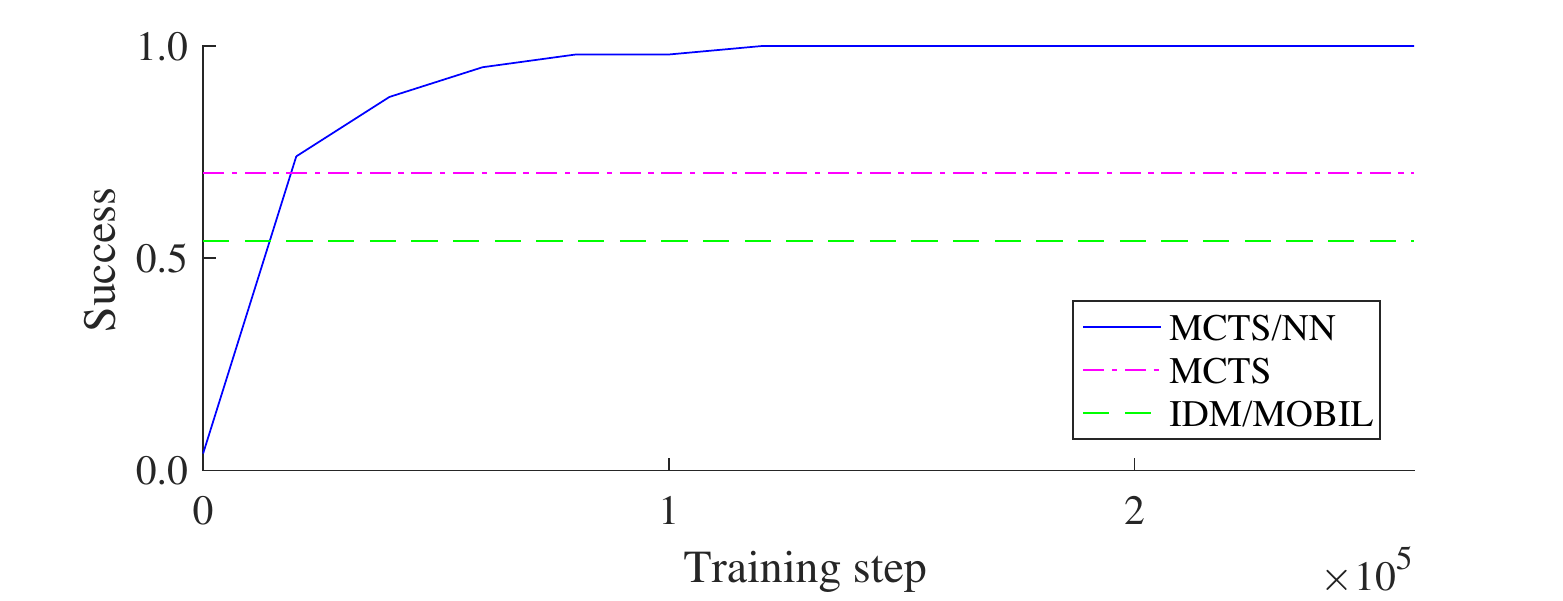}
        \caption{Proportion of successful evaluation episodes, as a function of training steps, for the highway exit case.}
    \label{fig:exitSuccess}
\end{figure}

\begin{figure}[!t]
    \centering
        \includegraphics[width=0.99\columnwidth]{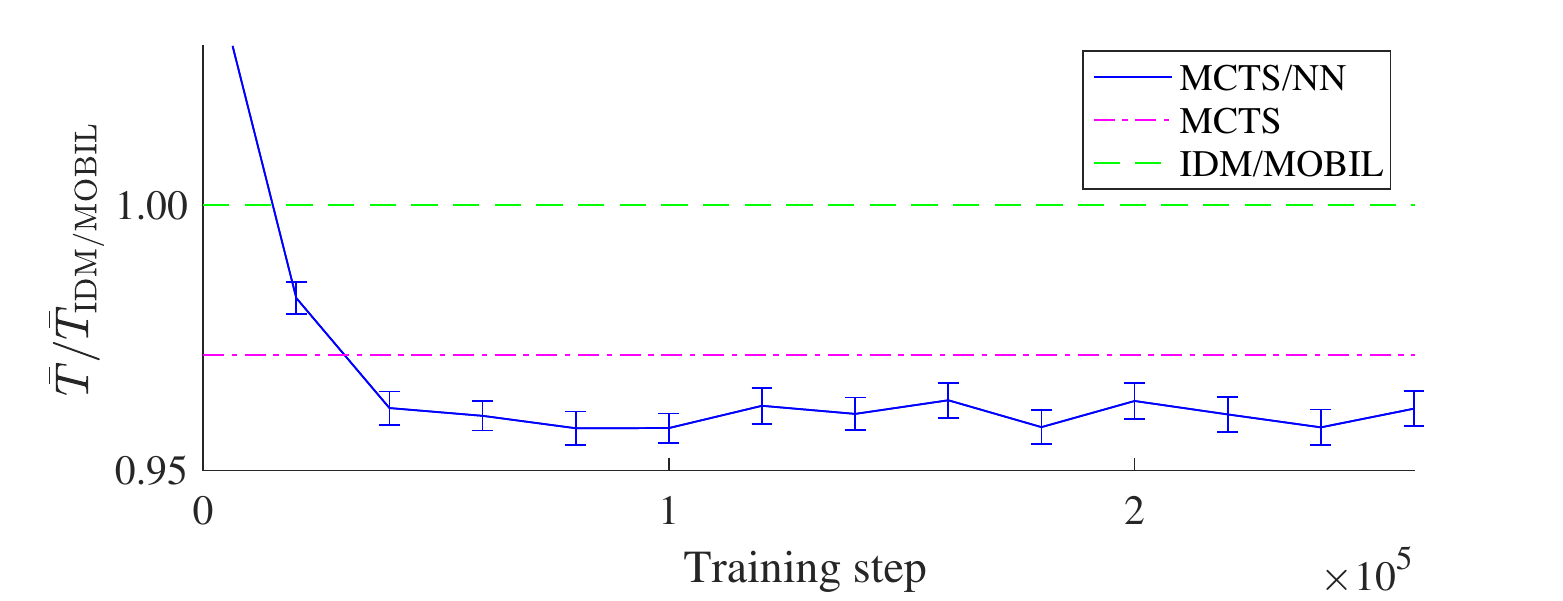}
        \caption{The average time $\bar{T}$ it took to reach the exit during the evaluation episodes, normalized with the time of the IDM/MOBIL agent $\bar{T}_\mathrm{IDM/MOBIL}$, as a function of training steps.
        The error bars indicate the standard error of the mean.}
    \label{fig:exitTime}
\end{figure}

The highway exit case is conceptually different from the continuous driving case, since it has a pass/fail outcome. In this case, the most important objective is to reach the exit, whereas a secondary objective is to do so in a time efficient way. Fig.~\ref{fig:exitSuccess} shows the proportion of evaluation episodes where the exit was reached, as a function of training steps for the MCTS/NN agent. The agent quickly learned how to reach the exit in most episodes, and after around $120{,}000$ training steps, it solved all of them. The success rate of the baseline methods was $70\%$ for the standard MCTS agent and $54\%$ for the modified IDM/MOBIL agent.
The time it took to reach the exit for the different methods is shown in Fig.~\ref{fig:exitTime}. Only the episodes where all methods succeeded to reach the exit are included in the comparison.

\begin{figure}[!t]
    \centering
        \subfloat[Initial state]{\includegraphics[width=0.45\columnwidth]{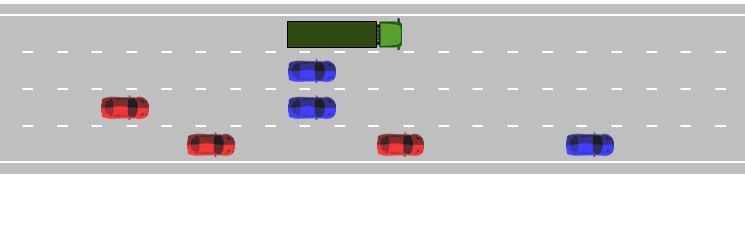}%
        \label{fig:exitStart}}
        \hfil
        \subfloat[At exit, IDM/MOBIL]{\includegraphics[width=0.45\columnwidth]{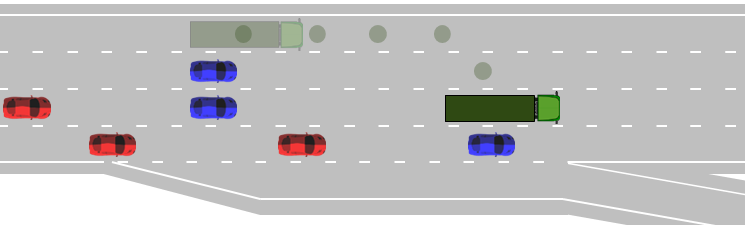}%
        \label{fig:exitRef}}
        \hfil
        \subfloat[At exit, MCTS]{\includegraphics[width=0.45\columnwidth]{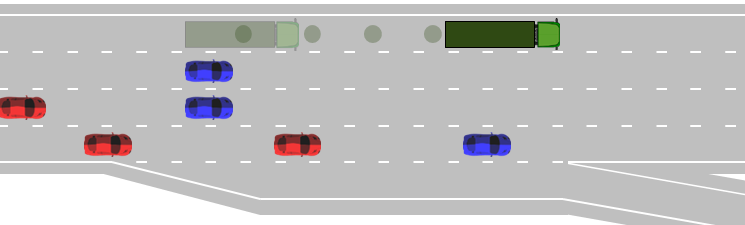}%
        \label{fig:exitMCTS}}
        \hfil
        \subfloat[At exit, MCTS/NN]{\includegraphics[width=0.45\columnwidth]{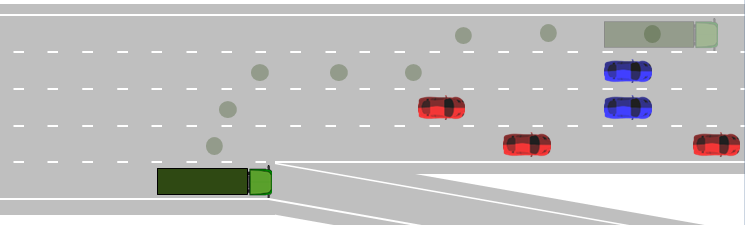}%
        \label{fig:exitMCTSNN}}
    \caption{Example of when it is necessary to plan relatively far into the future to solve a specific situation.
    The initial state is shown in (a) and the state at the exit is shown for the three agents in (b), (c), and (d). The dots show the position of the ego vehicle relative to the other vehicles during the maneuver, i.e., in (b) and (c) the ego vehicle accelerates and overtakes the slower vehicles, whereas in (d), the ego vehicle slows down and ends up behind the same vehicles.}
    \label{fig:exitSpecialCase}
\end{figure}

Similarly to the continuous highway driving case, there are situations for the highway exit case where it is necessary to plan over a long time horizon. One such situation is shown in Fig.~\ref{fig:exitSpecialCase}. There, the ego vehicle starts in the leftmost lane, $300$ m from the exit, and six vehicles are positioned in the other lanes. Three of the vehicles have timid driver parameters and the other three have aggressive driver parameters, except for the set speed, which is $v_\mathrm{set}=21$ m/s for all of them. All vehicles also start with an initial speed of $21$ m/s. The single way the ego vehicle can reach the exit in this situation is to first slow down and then make several lane changes to the right, which was only discovered by the trained MCTS/NN agent. The standard MCTS agent never found the way to the exit in its tree search and therefore stayed in its original lane, to not get negative rewards from changing lanes. The IDM/MOBIL agent accelerated to $25$ m/s and changed lanes to right as far as it could, without reaching the exit.

\begin{figure}[!t]
    \centering
        \includegraphics[width=0.99\columnwidth]{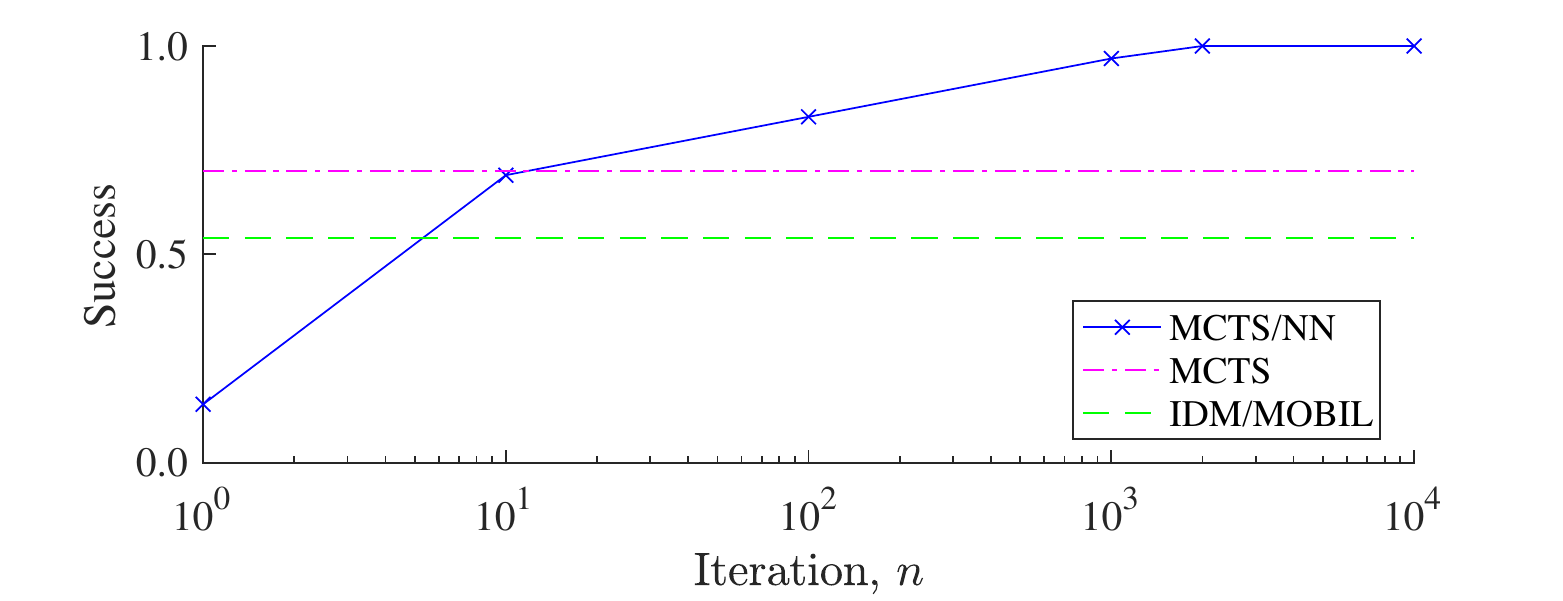}
        \caption{Proportion of successful episodes as a function of the number of MCTS iterations $n$ for the trained MCTS/NN agent, in the highway exit case.}
    \label{fig:exitSuccessSearches}
\end{figure}

The effect of the number of MCTS iterations $n$ is shown in Fig.~\ref{fig:exitSuccessSearches}. When using the learned policy from the neural network, i.e., one iteration, the MCTS/NN agent only succeeded in $14\%$ of the evaluation episodes. With ten iterations, the success rate matched the standard MCTS agent, which used $2{,}000$ iterations. Then, when the MCTS/NN agent also used $2{,}000$ iterations, it managed to plan far enough to solve all the evaluation episodes.

\begin{figure*}[!t]
    \begin{center}
        \includegraphics[width=1.99\columnwidth]{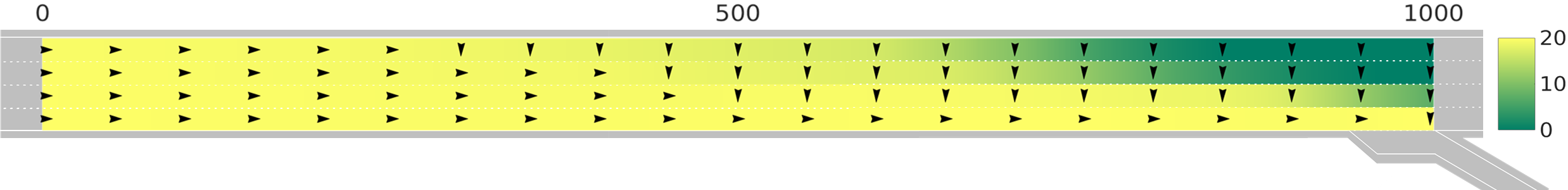}
    \end{center}
    \vspace{-15pt}
    \parbox{1.73\columnwidth}{\caption{This figure displays the learned values of different states $V(s,\theta)$ when there were no other vehicles present, for the highway exit case. The arrows represent which action that was taken for different states. An arrow that points to the right corresponds to action $a_1$, whereas downwards corresponds to $a_4$ (Table~\ref{tab:actions}). Note that the axes do not have the same scale.}
    \label{fig:exitValueMap}
    }
\end{figure*}

In order to illustrate the behavior of the trained MCTS/NN agent, Fig.~\ref{fig:exitValueMap} shows the learned value and the action that was taken for different states when approaching the exit, with no other vehicles present. The ego vehicle had an initial speed of $25$ m/s, $v_\mathrm{set}=25$ m/s, and $T_\mathrm{set}=2.5$ s. For states longitudinally far away from the exit, the agent learned that the value is $20$, which corresponds to expecting a reward of $1$ for all future steps (since the geometric sum of Eq.~\ref{eq:discountedReturn} equals $20$ for $\gamma=0.95$). As expected, the learned value decreases for all the lanes, except for the rightmost lane, for states close to the exit. Far from the exit, the agent always chooses action $a_1$, i.e., to stay in the current lane and keep its current speed, whereas closer to the exit, the agent changes lanes to the right, to bring it to the rightmost lane.

\section{Discussion}
\label{sec:discussion}

The results show that the agents that were obtained by applying the proposed framework to the two test cases outperform the baseline methods, and that the advantage is more significant for the highway exit case, especially in the number of solved test episodes (Fig.~\ref{fig:exitSuccess}). The reason for that the difference is larger in the exit case is that it is a more complex case, where the effect of the policy is more decisive. A suboptimal action in the continuous highway driving case just means a small time loss, whereas a mistake in the exit case can result in that the exit is not reached. Moreover, in general, the exit case requires a longer planning horizon than the continuous case, which is exemplified in Fig.~\ref{fig:exitSpecialCase}. In that situation, the standard MCTS did not reach deep enough in the search tree to figure out how it could reach the exit. However, the prior action probabilities and the value estimate of different states, which the MCTS/NN agent obtained from the training, allowed it to focus the tree search to the most promising regions. Thus, for the same number of MCTS iterations, it could search deeper in the tree and perform planning over a longer time horizon than the standard MCTS. In the example situation, the MCTS/NN agent was therefore able to figure out which actions that were needed in order to reach the exit.

The MCTS/NN agent is anytime capable, i.e., it can abort its search after any number of iterations, even after just one, which will then return the action given by the neural network. More iterations will in general improve the result, up to a limit, where the performance settles. In the cases considered in this study, full performance was reached at around $n=1{,}000$, see Fig.~\ref{fig:contPerformanceSearches} and~\ref{fig:exitSuccessSearches}. The number of searches that are necessary depends on the complexity of the environment and the specific traffic situation, which will require different planning depths, as was discussed above.

As mentioned in Sect.~\ref{sec:pomdpFormulation}, a simple reward model was used, which promotes driving close to the desired speed and penalizes lane changes. This model proved to work well in this study, but a more careful design may be required for other cases. Additional aspects, such as fuel efficiency or the influence on the surrounding traffic, could also be included. A reward model that mimics human preferences could be determined by using inverse reinforcement learning~\cite{NG2000}.

In a previous paper~\cite{Paper2}, we introduced a different method, where a DQN agent learned to make tactical decisions for a case that was similar to the continuous highway driving case described here. That method required around one order of magnitude more training samples to reach a similar performance as the MCTS/NN agent, which indicates the value of letting a planning component guide the learning process, from a sample efficiency perspective. However, each training sample is more computationally expensive to obtain for the method presented here, due to the many MCTS iterations that are done for each decision. If the training is carried out in a simulated environment where training samples are cheap, the advantage of the sample efficiency of the MCTS/NN agent can be argued, but if the training samples are obtained from real world driving where each sample is expensive, sample efficiency is important. For the two test cases in this study, the MCTS agent required around $100{,}000$ training samples, which corresponds to around $20$ hours of driving.

The generality of the proposed decision making framework was demonstrated by applying it to two different cases,
that are conceptually different. In the continuous highway driving case, the only goal is to navigate in traffic as efficiently as possible, whereas in the exit case, there is a terminal state with a pass/fail outcome, i.e., if the exit was reached or not.
In order to apply the framework to other cases, the following components need to be defined: the state space $\mathcal{S}$, the action space $\mathcal{A}$, the reward model $R$, the generative model $G$, and the belief state estimator.
However, the tree search and training process of Algorithm~\ref{alg:treeSearch} and~\ref{alg:training} would be identical for all cases.

When training a decision making agent by using the framework presented in this paper, or any other machine learning technique, it is important to note that the agent will only be able to solve the type of situations it encounters during the training process. Therefore, it is crucial to set up the training episodes so that they cover the intended case. Moreover, when using machine learning to create a decision making agent, it is difficult to guarantee functional safety of the agent's decisions. A common way to avoid this problem is to use an underlying safety layer, which ensures that the planned trajectory is safe before it is executed by the vehicle control system~\cite{Underwood2016}.

In this paper, the AlphaGo Zero algorithm was extended to a domain with a continuous state space, a not directly observable state, and where self-play cannot be used.
A generative model replaced the self-play component, progressive widening was added to deal with the continuous state space, and a state estimation component was added to handle the unknown state.
Furthermore, the CNN architecture of AlphaGo Zero, which due to Go's grid-like structure can be used to extract important features, was replaced by a CNN architecture that was applied to features of the surrounding vehicles.
One technical difference to the AlphaGo Zero algorithm is the UCB condition in Eq.~\ref{eq:ucb}, which determines which action to expand in the tree search. The numerator is here changed from AlphaGo Zero's $\sqrt{\sum_b{N(s,b)}}$ to $\sqrt{\sum_b{N(s,b)}+1}$, which means that when the tree search reaches a leaf node, it will choose to expand the action that is recommended by the neural network policy, i.e., $a = \argmax_a P(s,a,\theta)$, instead of a random action. This proved to be beneficial in this study, but more investigations are necessary to determine if it is beneficial in general.
Two other small technical differences are that the $Q$-value in the UCB condition is normalized, in order to keep $c_\mathrm{puct}$ constant for different environments, and that the $Q$-value of a leaf node is initialized to the value that is estimated by the neural network $V(s,\theta)$, which for this domain is a better estimate than setting it to zero, as in AlphaGo Zero.


\section{Conclusions}
\label{sec:conclusion}

The results of this paper show that the presented framework, which combines planning and learning, can be used to create a tactical decision making agent for autonomous driving. For two conceptually different highway driving cases, the resulting agent performed better than individually using planning, in the form of MCTS, or learning, in the form of the trained neural network. The agent also outperformed a baseline method, based on the IDM and MOBIL model. The presented framework is flexible and can easily be adapted to other driving environments. It is also sample efficient and required one order of magnitude less training samples than a DQN agent that was applied to a similar case~\cite{Paper2}.

\ifCLASSOPTIONcaptionsoff
  \newpage
\fi



%



\bibliography{references}{}
\bibliographystyle{ieeetr}

%

\begin{IEEEbiography}[{\includegraphics[width=1in,height=1.25in,clip,keepaspectratio]{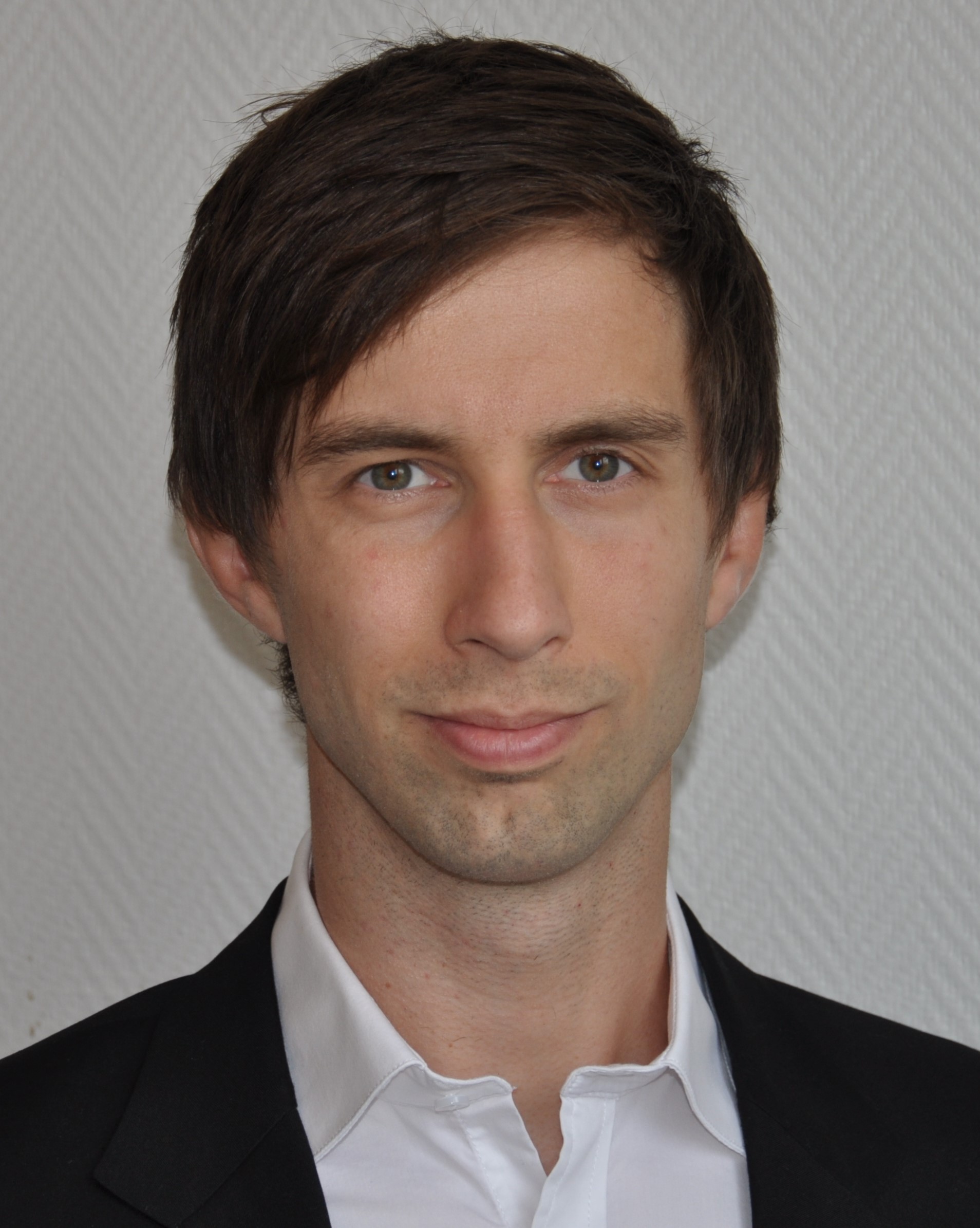}}]{Carl-Johan Hoel}
Carl-Johan Hoel received the B.S. and M.S degrees in physics from Chalmers University of Technology, Gothenburg, Sweden, in 2011. He is currently working towards the Ph.D. degree at Chalmers and Volvo Group, Gothenburg, Sweden. His research interests include reinforcement learning methods to create a general tactical decision making agent for autonomous driving.
\end{IEEEbiography}

\vspace*{\fill}

\begin{IEEEbiography}[{\includegraphics[width=1in,height=1.25in,clip,keepaspectratio]{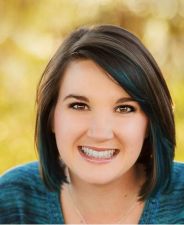}}]{Katherine Driggs-Campbell}
Katie Driggs-Campbell received her BSE in Electrical Engineering from Arizona State University and her MS and PhD in Electrical Engineering and Computer Science from the University of California, Berkeley.  She is currently an Assistant Professor in the ECE Department at the University of Illinois at Urbana-Champaign.  Her research focuses on exploring and uncovering structure in complex human-robot systems to create more intelligent, interactive autonomy. She draws from the fields of optimization, learning \& AI, and control theory, applied to human robot interaction and autonomous vehicles. 
\end{IEEEbiography}

\begin{IEEEbiography}[{\includegraphics[width=1in,height=1.25in,clip]{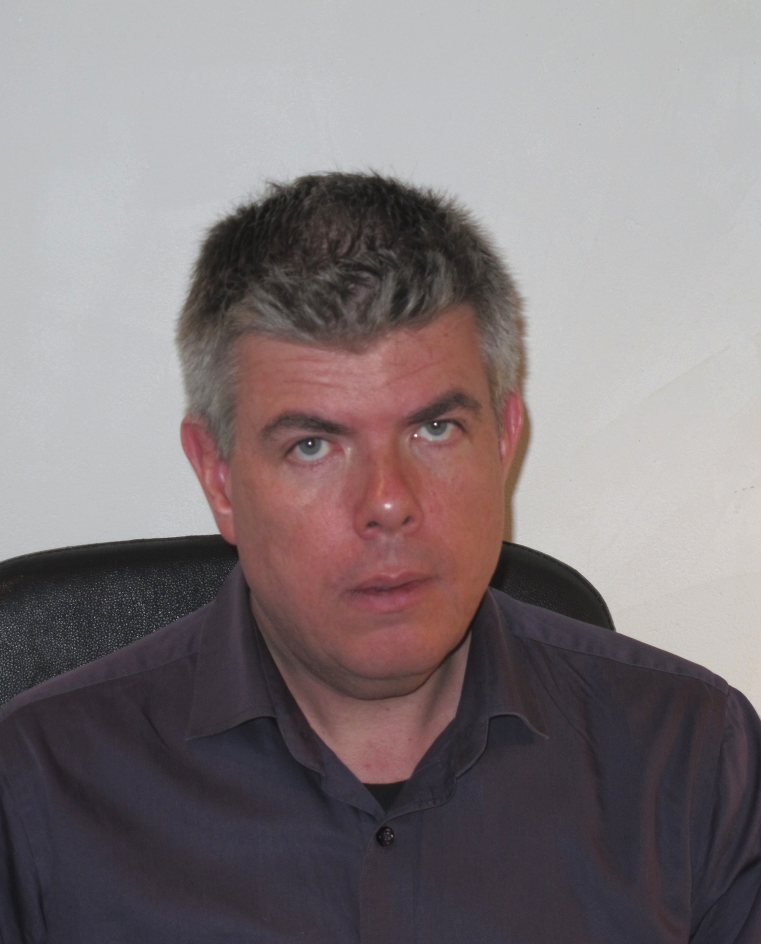}}]{Krister Wolff}
Krister Wolff received the M.S. degree in physics from Gothenburg University, Gothenburg, Sweden, and the Ph.D. degree from Chalmers University of Technology, Gothenburg, Sweden. He is currently an Associate Professor of adaptive systems, and he is also the Vice head of Department at Mechanics and maritime sciences, Chalmers. His research is within the application of AI in different domains, such as autonomous robots and self-driving vehicles, using machine learning and bio-inspired computational methods as the main approaches. 
\end{IEEEbiography}

\begin{IEEEbiography}[{\includegraphics[width=1in,height=1.25in,clip]{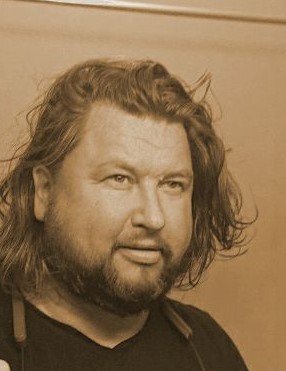}}]{Leo Laine}
Leo Laine received the Ph.D. degree from Chalmers
University of Technology, Gothenburg, Sweden,
within Vehicle Motion management. Since 2007, he has been 
with the Volvo Group Trucks Technology in the Vehicle 
Automation department. Since 2013, he has also been an Adjunct Professor in vehicle dynamics with Chalmers Vehicle Engineering and Autonomous Systems. Since 2013, he is specialist within complete vehicle control.
Since 2017, he is technical advisor within Vehicle Automation department.

\end{IEEEbiography}

\begin{IEEEbiography}[{\includegraphics[width=1in,height=1.25in,clip,keepaspectratio]{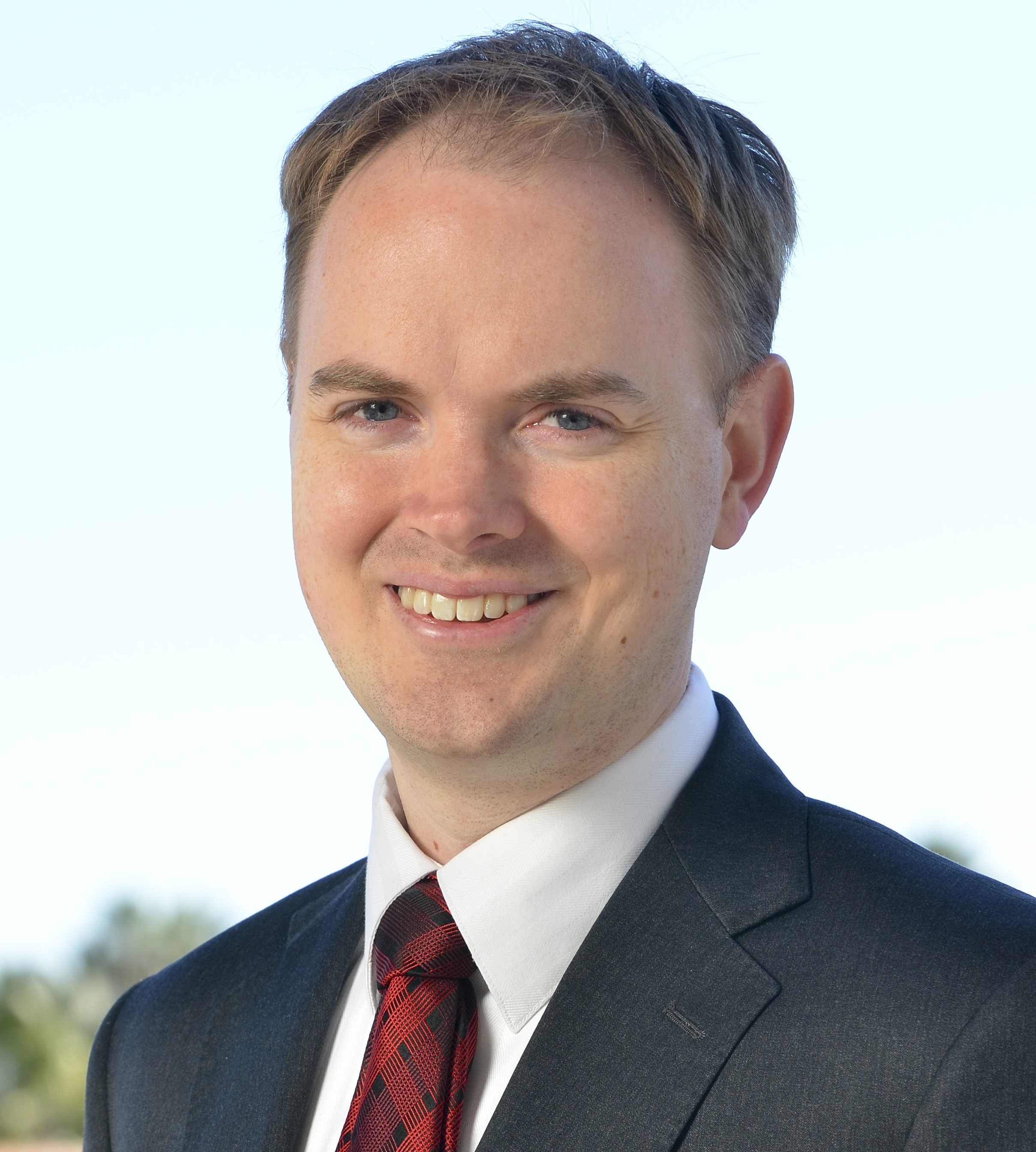}}]{Mykel J. Kochenderfer}
Mykel J. Kochenderfer received the B.S. and M.S. degrees in computer science from Stanford University, Stanford, CA, USA, and the Ph.D. degree from The University of Edinburgh, Edinburgh, U.K. He is currently an Assistant Professor of aeronautics and astronautics with Stanford University. He is also the Director of the Stanford Intelligent Systems Laboratory, conducting research on advanced algorithms and analytical methods for the design of robust decision-making systems.
\end{IEEEbiography}







\enlargethispage{-1.6in}

\end{document}